\documentclass[a4paper,fleqn]{cas-dc}

\usepackage[authoryear,longnamesfirst]{natbib}

\usepackage{algorithm}
\usepackage{algpseudocode}
\usepackage{amsmath}
\usepackage{subcaption}
\usepackage{graphicx}

\newcommand{\ml}[1]{}

\def\tsc#1{\csdef{#1}{\textsc{\lowercase{#1}}\xspace}}
\tsc{WGM}
\tsc{QE}
\tsc{EP}
\tsc{PMS}
\tsc{BEC}
\tsc{DE}

\begin{document}
\let\WriteBookmarks\relax
\def\floatpagepagefraction{1}
\def\textpagefraction{.001}

\shorttitle{Compositional Fusion of Signals in Data Embedding}

\shortauthors{Guo, Xu, Lewis, and Cristianini}

\title [mode = title]{Compositional Fusion of Signals in Data Embedding }                  
\tnotemark[1,2]
\author{Zhijin Guo}
\author{Zhaozhen Xu}
\author{Martha Lewis}

\author%
{Nello Cristianini}

\begin{abstract}
Embeddings in AI convert symbolic structures into fixed-dimensional vectors, effectively fusing multiple signals. However, the nature of this fusion in real-world data is often unclear. To address this, we introduce two methods: (1) Correlation-based Fusion Detection, measuring correlation between known attributes and embeddings, and (2) Additive Fusion Detection, viewing embeddings as sums of individual vectors representing attributes.

Applying these methods, word embeddings were found to combine semantic and morphological signals. BERT sentence embeddings were decomposed into individual word vectors of subject, verb and object. In the knowledge graph-based recommender system, user embeddings, even without training on demographic data, exhibited signals of demographics like age and gender.

This study highlights that embeddings are fusions of multiple signals, from Word2Vec components to demographic hints in graph embeddings.

\end{abstract}


\maketitle
\section{Introduction}
\label{sec: introduction}

In AI research, embeddings are used to represent symbolic structures such as 
\emph{knowledge graphs} as collections of vectors of fixed dimension. By converting to embeddings, standard algebraic techniques can be used to perform inferences on symbolic data. In other words, using embeddings allows for a convenient way to model and process data. This paper examines the extent to which vector embeddings can be viewed as a fusion of informative signals are encoded in embeddings, and how those signals can be disentangled and interpreted.

Knowledge graphs are a way of encoding explicit declarative knowledge about a set of entities in a domain and the relations between those entities. They are a powerful tool to capture structured information about the world and model complex relationships between various entities. With the rise of massive knowledge bases and the need for efficient querying and inference, traditional symbolic reasoning on knowledge graphs can become computationally expensive.

To address these challenges, \emph{graph embeddings} have been introduced as a method to convert the structured information of knowledge graphs into a continuous vector space. These embeddings aim to capture the topological relations and semantic meanings of entities and relationships in the graph. The conversion of symbolic constructs such as knowledge graphs into continuous embeddings enables efficient algebraic operations, similarity calculations, and other tasks. For instance, in bipartite graph representations, graph embeddings can reflect properties like a user liking a certain movie. The efficiency and expressiveness of these embeddings have proven useful across many applications, including link prediction (which we focus on here), node classification \citep{ji2021survey}, and graph generation \citep{bo2021social}.

Many problems can naturally be cast in a knowledge graph setting, by defining the entitites and the relation(s) between them. For example, the standard technique known as \emph{word embedding} defines the entities as words, and the relation between words as one of “co-occurrence”, such that two words are related if they often occur in the vicinity of one another. In this and in many other cases, the strength of the relation is used too, and can be represented as a weight on the edge of the graph.

\subsection*{Problem}
However, a challenge arises: these embeddings, drawn from real-world data to encode either graph topological or word context relations, may not always be transparent to human interpretation. Attempting to interpret embeddings in a compositional way implies that an embedding can be viewed as a \textbf{fusion} of distinct information components. However, this opacity makes potential unintended information hard to detect and assess, further complicating our understanding of how different components merge within the embedding space.

\subsection*{On Compositionality}
In word embeddings, a series of interesting phenomena have been noted, whose extension to other forms of data is of great practical interest. They include “compositionality”, that is, the property that the embedding of two words that have certain semantic or syntactic relations are related in a predictable manner, typically in an additive form. This allows for certain types of inference to be performed. A classic illustration \citep{mikolov2013distributed} is the relationship between the embeddings of the words "King" and "Queen":
$$\mathbf{x}_{king}-\mathbf{x}_{man}+\mathbf{x}_{woman} \approx \mathbf{x}_{queen}$$
This provides the possibility to perform analogical inferences, where we can predict relationships (such as gender) between words. 

Both the phenomena of compositionality and of bias in embeddings can be traced back to the \emph{distributional hypothesis} \citep{harris1954distributional}. This posits that words that frequently appear in similar contexts tend to have related meanings. For instance, "doctor" and "nurse" often co-occur with terms like "hospital" and "patient", hence their embeddings will be close, indicating semantic similarity.. While this assumption is powerful for capturing semantic relationships and nuances, it also means that any biases present in the data – stemming from societal norms, customs, or even data collection methods – get encoded into the embeddings.
\subsection*{Approaches to Understanding Compositionality in Embeddings}
One major unresolved concern in word embedding is whether compositionality emerges spontaneously from distributional semantics or is an inherent feature \citep{mikolov2013distributed}. While the concept of compositionality was originally rooted in linguistics, its application to vector embeddings—replacing string concatenation with vector addition—has raised questions about its practicality and significance in the realm of computational linguistics \citep{andreas2019measuring}.

Moreover, while machine learning techniques like Disentangled Representation Learning (DRL) aim to address these gaps by segmenting attributes within data representations \citep{bengio2013representation}, \citet{shwartz2019still} undertook an examination of word representation compositionality via six tasks, probing into the phenomena of semantic drift and implicit meaning. \citet{andreas2019measuring} postulated a metric for compositionality based on the approximation fidelity of observed representations when assembled from inferred primitives. This scholar also introduced the Tree Reconstruction Error (TRE) method, focused on gauging the compositionality through multiplication. \citet{murty2022characterizing} found that, when trained on language tasks, increasingly adopt a hierarchical, tree-like processing approach, which improves their compositional generalization capabilities.

While the concept of compositionality has been deeply studied in fields like linguistics, most of their works primarily focus on language. On the other hand, there is a lack of tools that can measure the degree of compositional structure in vector representations.

Issues in Sentence Embedding Decomposition:  BERT \citep{devlin2018bert} learns significant syntactic information without explicit syntactic trees during its training \citep{hewitt2019structural}. \citet{ettinger2016probing} created a dataset for identifying semantic roles in embeddings and examined altered sentence meanings with minimal lexical changes, but did not address how these embeddings understood broader language nuances. \citet{dasgupta2018evaluating} made a dataset for word combination studies in embeddings, emphasizing changes in word order and specific word additions, but it is unclear how these modifications affect overall sentence understanding. While \citet{adi2016fine} introduced techniques to evaluate sentence embeddings, such as measuring sentence length and determining word order, and found LSTM auto-encoders effective, their approach did not differentiate between word and sentence embeddings, leaving the relationship between individual word representations and their corresponding sentence embedding unexplored.

Algorithmic Bias in Graph Embedding: As the application of embeddings expands, concerns over biases in machine learning emerge \citep{bolukbasi2016man}. Biases in data embeddings can inadvertently reflect societal norms and prejudices. For instance, associations in word embeddings often reveal embedded gender biases \citep{jonauskaite2021english,sutton2018biased, caliskan2017semantics}. This algorithmic bias could manifest in various machine learning applications, requiring proactive detection and mitigation methods, as argued by \citet{fisher-etal-2020-debiasing}.

\paragraph{Our methods}
Our work is most aligned with that of \citet{andreas2019measuring, hewitt2019structural, bose2019compositional}. We are interested in the extent to which embeddings can be additively decomposed into component parts. We examine three different kinds of data embedding: 1) word embeddings, 2) sentence embeddings, and 3) knowledge graph embeddings.

In the example of word embedding, we use pretrained Word2vec \citep{mikolov2013efficient} embeddings and investigate the extent to which these word embeddings can be analysed as a fusion of their semantic meaning and their syntactic structure. In the example of sentence embeddings, we use sentence embeddings from  BERT \citep{devlin2018bert}, and look at the extent to which simple sentences may be analysed as an additive fusion of their constituent words. Finally, we look at knowledge graph embedding.  In this problem, we train a set of embeddings over the MovieLens dataset \citep{harper2015movielens}. This dataset contains entities for users and entities for movies, and relations on the knowledge graph consist of the users' ratings of the movies. We train our embeddings with the objective of performing \emph{link prediction}, that is, the task of predicting whether a link holds between two entities. We describe this in more detail in section \ref{sec: section2}, however, the key point is that we learn the embeddings \emph{without any reference to the demographic attributes of the users}, e.g. gender or age. We investigate the extent to which the user embeddings are in fact composed of an additive fusion of demographic attributes, even though these are not used in training.

Throughout the three problems described above, we ask whether we can decompose an embedding into interpretable components. Specifically, we investigate additive  decompositions, that is of the type $\phi(x) = \phi(x1) +\phi(x2)$.

\begin{figure}[htbp]
         \centering
         \includegraphics[scale=.40]{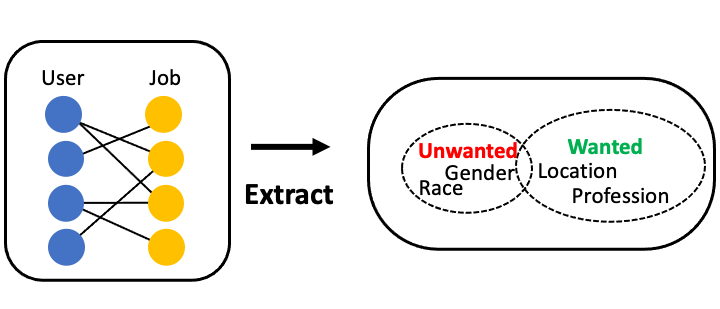}
         \caption{Embedding contains an information fusion of both wanted and unwanted information}
          \label{fig: unwated info}
\end{figure}

\subsection*{Our Approach }

We introduce two distinct methods to analyse the extent to which embeddings of can be interpreted as a fusion of interpretable components.

\begin{enumerate}
    \item \textbf{Correlation-based Fusion Detection} We use Canonical Correlation Analysis (CCA) to provide a novel approach to measure the correlation between interpretable attributes and the data embedding itself. This method provides a quantitative measure of compositionality.

    \item \textbf{Additive Fusion Detection} We treat embeddings as additive fusions of meaningful vector directions. We view an embedding \( v \) as an aggregated sum \( v = x_{1} + x_{2} + \ldots + x_{k} \), with each component \( x_{i} \) a distinct meaningful direction within the vector space that represents an attribute (such as gender, age, etc.).
\end{enumerate}

\subsection*{Improvements Over Previous Approaches}

Unlike earlier models, our methods are versatile across different embedding types. Approaches such as \citet{shwartz2019still} \citet{mikolov2013distributed} consider only how word embeddings should be decomposed. Similarly, \citet{bose2019compositional} \citet{fisher-etal-2020-debiasing} consider only the interpretation of graph embeddings. Here, we show that the same methods can be used across different embedding types. 

While \citet{mikolov2013distributed, bose2019compositional} show that embeddings can be decomposed into simple attributes, they only provide a qualitative decomposition, whereas we are able to provide a weighting that quantifies how much each component contributes to the overall fusion of attributes by the correlation-based fusion detection. 

Furthermore, our Additive-Fusion Detection method provides a novel way to detect signal fusion in embeddings. We consider an embedding \( v \) as a cumulative sum given by \( v = x_{1} + x_{2} + \dots + x_{k} \), where each \( x_{i} \) denotes a unique direction in the vector space corresponding to attributes. This was already done implicitly by \citet{mikolov2013distributed}, however, we provide a systematic method by which to isolate signals in the vector space and confirm the robustness of these signals via statistical testing.

\subsection*{Relation to Information Fusion}

Our approach is deeply rooted in information fusion. By treating embeddings as additive composites of discrete, meaningful vector directions, we are essentially fusing information from various attributes. This fusion offers a more cohesive understanding and enhanced interpretability of embeddings. Whether it is the cumulative representation of a sentence via its grammatical components or a user's demographic description, our methods demonstrate the power of information fusion in understanding and improving embeddings.

\subsection*{Results}
We apply our methods to word embeddings, sentence embeddings, and graph embeddings. We find that word embeddings can be decomposed into semantic and morphological components. Similarly, for BERT sentence embeddings, we find that the sentence embeddings can be decomposed into a sum of individual word embeddings. Finally, we show that embeddings corresponding to users in a database of users and movie ratings can be decomposed into a sum of embeddings corresponding to demographic attributes such as gender, age, and so on, \emph{even though these attributes are not used in the training of the embeddings}.

Our findings significantly advance the understanding of embeddings. In word embeddings, we revealed the multidimensional richness within Word2Vec, highlighting opportunities for detailed analysis, from semantics to morphology. Our decomposition techniques in sentence embeddings showed that BERT's embeddings can be decomposed into the contributions of the subject, verb and object. Most crucially, in graph embeddings, we discerned that user embeddings capture private demographic attributes, illustrated by the ability to compute composite embeddings like that of a "50-year-old female" from individual attribute embeddings. This insight into detecting private attributes in systems, like movies, is pivotal for future research.

\subsection*{Structure of Paper}
Section \ref{sec: section2} covers embedding, mapping elements to vector spaces, focusing on word, sentence, and graph embeddings. Furthermore, we discuss how the linguistics idea of compositionality applies to the fusion of different signals in vector embeddings. Section \ref{sec: method} introduces two methods: \emph{Correlation-based Fusion Detection} and \emph{Additive Fusion Detection} to detect the fusion of signals in data embeddings. Section \ref{sec: experiments} presents experiments on three data embeddings, and Section 5 discusses results.

\begin{figure*}[htbp]
         \centering
         \includegraphics[scale=.32]{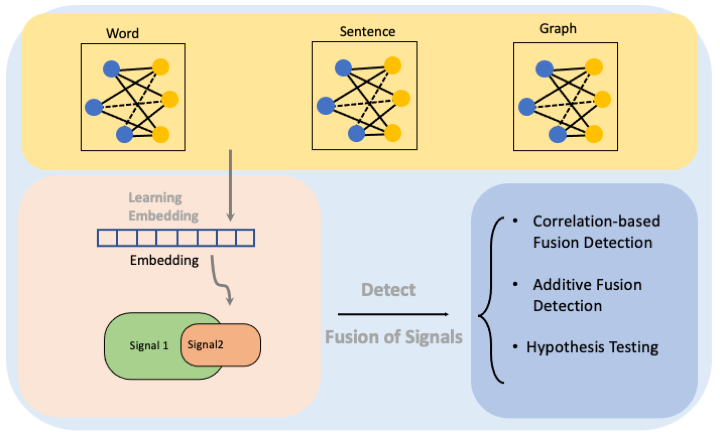}
         \caption{Structure of the paper}
          \label{fig:paper-structure}
\end{figure*}

\section{Background and Related Work}
\label{sec: section2}
\subsection{Embedding}
 In machine learning, embedding is the process of mapping elements from a set, denoted as $I$, to points in a vector space. We write a set of coordinates \textbf{B} to represent the items of $I$ as follows: $$\textbf{B}=\Phi(I)$$ 
where $\Phi$ is the mapping function that maps the items (elements of the set) to their coordinates. This embedding function can be learned from a set of data containing those items: for words, this can be done by exploiting co-occurrence statistics between words; for elements of a graph, by exploiting the topology, i.e., the relations between different elements.  

More generally, we can consider any kernel-based method as an example of embedding, since it depends on defining a \emph{kernel function} that generates a \emph{kernel matrix} once applied to the set of items, and this one can be regarded as an \emph{inner product matrix} in an embedding space (also known as the feature space).

Formally, for two data points \(x\) and \(y\), a kernel function \citep{shawe2004kernel} is defined as:
\[ K(x, y) = \langle \phi(x), \phi(y) \rangle \]
where \(\phi\) is a mapping from the input space to the feature space. The function \(K\) gives the inner product between the images of \(x\) and \(y\) in the feature space. However, the exact form of \(\phi\) doesn't need to be known as long as we can compute \(K\).

In this case, knowing the kernel (that is, the relation) between any two items is sufficient, and often the actual coordinates of the embedding are not known. We could also consider part of the same category any feature-based description of data: once a set of measurements is defined, they can be used to generate a vector that describes the item, which in turn can be regarded as coordinates (assuming those are numeric measurements). So an embedding is defined every time we agree on a set of measurable properties (features) or on a kernel function. 

In the example of word embeddings and knowledge graph embedding we will make use of co-occurrence or relational information to create the embedding. In the example of sentence embedding we will make use of a feature vector, as defined by a tool known as BERT. In both cases we will be interested how the embeddings of structured objects (e.g. sentences) can depend on the relations between those structures.

\subsubsection{Word Embedding} 
\paragraph{Word2Vec}
Word2Vec, as introduced by \citet{mikolov2013efficient}, is a method to embed words into vectors based on the distributional hypothesis: words in similar contexts have similar meanings. It consists of two architectures: Continuous Bag-of-Words (CBOW) and Skip-Gram. CBOW predicts a word from its context, while Skip-Gram predicts context words from a target word. 

Formally, for vectors of two words \(x\) and \(y\), their similarity in the embedded space can be computed as:
\[ K(x, y) = \langle x, y \rangle \]
This dot product serves as an effective metric for semantic similarity, capturing the relation of cooccurrence between words. While Word2Vec doesn't directly compute co-occurrence statistics, the embeddings inherently reflect these relations due to the optimization objectives.

\subsubsection{Sentence Embedding: BERT}
\label{sec:se}
We also consider the problem of deriving the meaning of sentences from the meaning of the words within them. We look at sentence embeddings extracted from BERT. 

BERT, introduced by \citet{devlin2018bert}, is a pre-trained Transformer-based model capturing bidirectional contexts of words, producing nuanced sentence embeddings. Unlike models like GloVe \citep{pennington-etal-2014-glove}, BERT doesn't use explicit co-occurrence statistics but learns context through deep training. The attention mechanisms within BERT employ dot products, serving as implicit kernel functions that dictate the relationship between parts of input text, reminiscent of the kernel function defined as:
\[ K(x, y) = \langle x, y \rangle \]
SBERT \citep{reimers2019sentence}, a sentence embedding derivative of BERT, was trained on natural language inference (NLI) corpora \citep{bowman2015large,williams2017broad}. .

\begin{figure}[hbtp]
    \centering
    \includegraphics[height=3in]{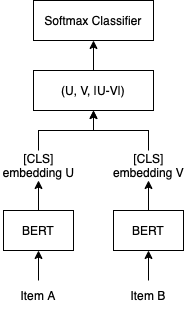}
    \caption{SBERT training process. Two BERTs in the graph are identical and share all the parameters. After BERTs generate the embeddings for input items, the embeddings are concatenated and classified with a softmax classifier. All the parameters in BERT and softmax classifier are updated during the training.}
    \label{fig:sbert}
\end{figure}

For each input token, BERT generates an output vector, where $\Phi_{BERT}: X \rightarrow Y \in\mathbb{R}^{768}$. The output vector of the [CLS] token is usually used for classification tasks because it can represent the information of the entire input sequence. However, the representation generated by pre-trained BERT fails to capture sentence similarity. Ideally, the sentence embeddings with similar meanings will be close to each other in the vector space. Thus, we use SBERT \citep{reimers2019sentence}, a version of BERT trained specifically for generating sentence representation that can be compared using cosine similarity. It created a leading performance on semantic textual similarity (STS) task \citep{cer2017semeval} by introducing a Siamese structure, as shown in figure \ref{fig:sbert}.

SBERT creates a state-of-the-art performance on variable STS tasks compared to existing sentence embeddings, such as InferSent \citep{conneau-EtAl:2017:EMNLP2017} and Universal Sentence Encoder \citep{cer2018universal}. Using SBERT to generate sentence embedding helps us look into BERT's mechanism while investigating the compositionality in the embedding.

\subsubsection{Knowledge Graph Embedding}
A \textit{graph} $G = (V, E)$ consists of a set of vertices $V$ with edges $E$ between pairs of vertices. In a \textit{knowledge graph}, the vertices $V$ represent entities in the real world, and the edges $E$ encode that some relation holds between a pair of vertices. As a running example, we consider the case where the vertices $V$ are a set of viewers and films, and the edges $E$ encode the fact that a viewer has rated a film.

Knowledge Graphs represent information in terms of entities (or nodes) and the relationships (or edges) between them. The specific relation \( r \) that exists between two entities is depicted as a directed edge, and this connection is represented by a triple \((h,r,t)\). In this structure, we distinguish between the two nodes involved: the \emph{head}  (\( h \)) and the \emph{tail}  (\( t \)), represented by vectors \( \mathbf{h} \) and \( \mathbf{t} \) respectively. Such a triple is termed a \emph{fact}, denoted by \( f \):
\[
f = (h,r,t)
\]

In order to mathematically capture the relationships and structures within a knowledge graph, we employ the concept of embeddings. A knowledge graph embedding assigns vectors to nodes and edges in such a way that the graph's topology is encoded. To be specific, a vector \( \mathbf{x} \in \mathbb{R}^n \) is allotted to each member of \( V \), ensuring the existence of a distance function \( D(\mathbf{x}_{i},\mathbf{x}_{j}) \) where  \( E(v_{i},v_{j})=1 \iff D(\mathbf{x}_{i},\mathbf{x}_{j}) < \theta \) for a certain threshold \( \theta \). We refer to these vectors \( \mathbf{x} \) as the \emph{embedding} of the nodes. The function that facilitates this embedding is the \emph{embedding function}: \( \Phi_{KG}: V \rightarrow \mathbb{R}^n \), or \( \mathbf{x}=\Phi(v) \).

Conversely, given a set of points in a space, we can link them to form a graph. The decision of which pairs of nodes $\langle v_{i}, v_{j}\rangle$ should be linked is made by using a scoring function $f(\mathbf{x}_{i}, \mathbf{x}_{j})$ that will be learnt from data. Unlike typical kernel methods which evaluate pairwise data, the Knowledge Graph Embedding's kernel operates on triplets, aligning with the relational architecture of knowledge graphs. Two commonly used functions generating a score between  $\mathbf{x}_{i}$ and $\mathbf{x}_{j}$ are:
\begin{align} \label{eq:mult} \textbf{Multiplicative: }&S(\mathbf{x_{i}}, \mathbf{x_{j}})= \mathbf{x_{i}}^{T} \mathbf{R} \mathbf{x_{j}} \\
\textbf{Additive: }&S(\mathbf{x_{i}}, \mathbf{x_{j}})=\|\mathbf{x_{i}}+\mathbf{r}-\mathbf{x_{j}}\|
\end{align}

where $\mathbf{R}$ and $\mathbf{r}$ are parameterised matrices or vectors that will be defined below. We can think of different $\mathbf{R}_i$ and $\mathbf{r}_i$ as encoding specific relations, allowing the same entity embedding $\mathbf{x}$ to participate in multiple different relations.

\begin{figure}
\centering
\begin{subfigure}{.25\textwidth}
  \centering
  \includegraphics[width=.8\linewidth]{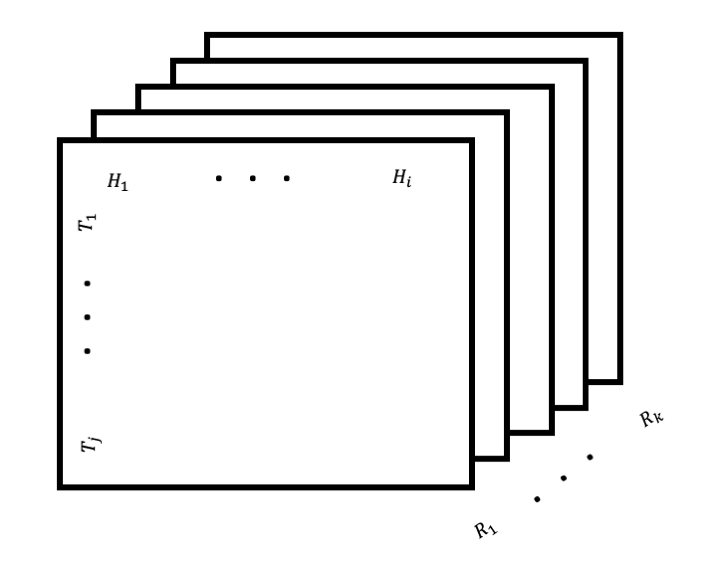}
  \caption{Multiplicative form \ml{the writing in this figure is too small}}
  \label{fig: frontal-slice}
\end{subfigure}%
\begin{subfigure}{.25\textwidth}
  \centering
  \includegraphics[width=.8\linewidth]{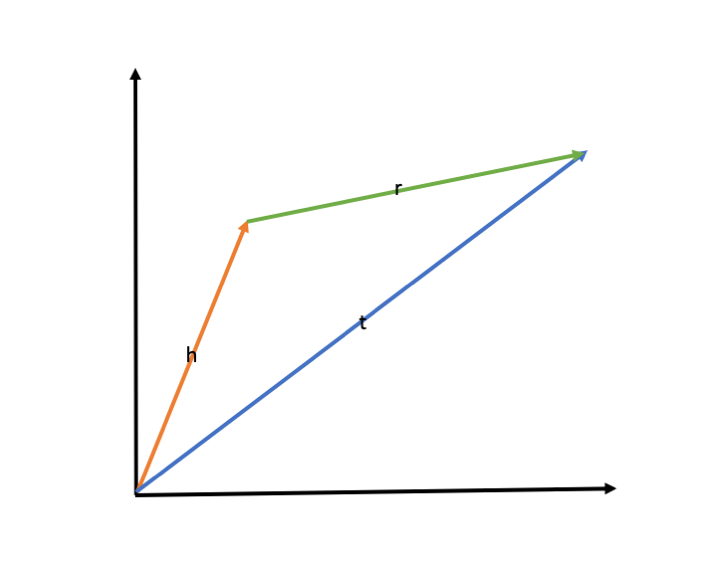}
  \caption{Additive form}
  \label{fig: transe}
\end{subfigure}
\caption{Two Different Scoring Functions}
\label{fig:test}
\end{figure}

We will follow this convention below, and use the multiplicative form of the scoring function which follows the settings of \citet{berg2017graph}
\paragraph{Multiplicative Scoring Function}
\citet{nickel2011three} proposed a tensor-factorisation based model for relational learning, in which they treat each frontal slice, as shown in Figure \ref{fig: frontal-slice}) of the tensor as a co-occurrence matrix for each entity with a given specific relation. Such a tensor could then be decomposed into three different tensors for the head entity, relation and tail entity.
For example, consider a 3D tensor, and we are looking at its frontal slices. The $i,j$ entry of the $k$-th frontal slice encodes the interaction between the head entity $h_{i}$, the relation $R_{k}$, and the tail entity $t_{j}$. This entry  can be decomposed into the product of $\mathbf{h_{i}}$, $\mathbf{R_{k}}$ and $\mathbf{t_{j}}$ A scoring function of a triple could also explain this in multiplicative way. We use $S(f)$ to denote the score of a triple $(h,r,t)$ and we use $\mathbf{h}, \mathbf{R}, \mathbf{t}$ (vectors) to denote the embeddings of each element of the triple $f = (h,r,t) \in F$.
\begin{equation} 
S(f)=\mathbf{h}^{T} \mathbf{R} \mathbf{t} \quad \quad \quad \mathbf{h} \in \mathbb{R}^{d}, \mathbf{R} \in \mathbb{R}^{d \times d}, \mathbf{t} \in \mathbb{R}^{d}
\end{equation}

Various model variations exist. DistMult \citep{yang2014embedding} retains only the $R$ matrix diagonal, reducing over-fitting. ComplEx \citep{trouillon2016complex} uses complex vectors for asymmetric relations. See Figure \ref{fig: frontal-slice} for a multiplicative scoring illustration.

In this work, we will be using DistMult \citep{yang2014embedding} for the models. DistMult is favored for its simplicity and computational efficiency, especially its adeptness at capturing symmetric relations using element-wise multiplication of entity embeddings, which also makes it scalable for large knowledge graphs.

\paragraph{Additive Scoring Function}
\citet{bordes2013translating} introduced TransE, where relationships translate entities in the embedding space. For instance, $\mathbf{h}(King)+\mathbf{r}(FemaleOf)\approx\mathbf{t}(Queen)$. Figure \ref{fig: transe} illustrates the additive scoring of this model.
\begin{equation}
    S(f)=\|\mathbf{h}+\mathbf{r}-\mathbf{t}\| \quad \quad\quad \mathbf{h} \in \mathbb{R}^{d}, \mathbf{r} \in \mathbb{R}^{d}, \mathbf{t} \in \mathbb{R}^{d}
    \label{transe}
\end{equation}

\paragraph{Rating Prediction}

In alignment with \citep{berg2017graph}, we establish a function $P$ that, given a triple of embeddings $(\mathbf{h},\mathbf{R},\mathbf{t})$, calculates the probability of the relation against all potential alternatives.
\begin{eqnarray}
P\left(\mathbf{h},\mathbf{R},\mathbf{t}\right)=\text{SoftArgmax}(S(f)) =\frac{e^{S(f)}}{e^{S(f)}+\sum_{r' \neq r\in \mathscr{R}} e^{S(f')}}
\label{eqn: loss}
\end{eqnarray}
In the above formula, $f =(h,r,t)$ denotes a true triple, and $f'=(h,r',t)$ denotes a corrupted triple, that is a randomly generated one, that we use as a proxy for a negative example (a pair of nodes that are not connected). 

Assigning numerical values to relations $r$, the predicted relation is then just the expected value $
\text{prediction} = \sum_{r \in \mathscr{R}} r P\left(\mathbf{h},\mathbf{R},\mathbf{t}\right)$
In our application of viewers and movies, the set of relations $\mathscr{R}$ could be the possible ratings that a user can give a movie. The predicted rating is then the expected value of the ratings, given the probability distribution produced by the scoring function. $S(f)$ refers to the scoring function in \citet{yang2014embedding}.

To learn a graph embedding, we follow the setting of \citet{bose2019compositional} as follows, 

\begin{equation}
        L  = - \sum_{f \in \mathscr{F}} \log \frac{e^{S(f)}}{e^{S(f)}+\sum_{f' \in \mathscr{F}'} e^{S(f')}}
     \label{eq:actual_loss}
\end{equation}
This loss function maximises the probabilities of true triples $(f)$ and minimises the probability of triples with corrupted triples: $(f')$.

\paragraph{Evaluation Metrics}
We use 4 metrics to evaluate our performance on the link prediction task. 
These are root mean square error (RMSE, $\sqrt{\frac{1}{n} \sum_{i=1}^{n}\left(\hat{y}_{i}-y_{i}\right)^{2}}$, where $\hat{y}_i$ is our predicted relation and $y_i$ is the true relation), Hits@K - the probability that our target value is in the top $K$ predictions, mean rank (MR) - the average ranking of each prediction, and mean reciprocal rank (MRR) to evaluate our performance on the link prediction task. These are standard metrics in the knowledge graph embedding community.

\subsection{Compositionality}
A property of certain embeddings that has the potential to help with the above concerns (as well as others) is that of “compositionality”. Introduced in the domain of traditional linguistics, this property has been extended to also cover vector representations. Traditionally it refers to how the meaning of a linguistic expression results from its components. For example, the word ``compositionality'' can be viewed as the concatenation of multiple parts ``Com+pos+ition+al+ity'' that modify the meaning of the initial word stem.

In the case of vector embeddings, we substitute the concatenation operation with the vector addition operation, so that a vector representation is compositional if it can be regarded as the sum of a small set of components (which can hopefully be interpreted and even manipulated). Introduced in the domain of traditional linguistics, this property has been extended to also cover vector representations. Traditionally it refers to how the meaning of a linguistic expression results from its components. 
For example, we could imagine an embedding $\Phi$ that maps from items (tokens) to vectors in such a way that
$$\Phi(compositionality) \approx \Phi(com) + \Phi(pos) + \Phi(ition)+ \Phi(ality)$$

\subsubsection{Compositionality in Word Embedding}

Recall the example \citep{mikolov2013distributed} in Section \ref{sec: introduction} involves the difference between how the words ``King" and `` Queen" are embedded as: $\mathbf{x}_{king}-\mathbf{x}_{man}+\mathbf{x}_{woman} \approx \mathbf{x}_{queen}$.

An interesting question is whether this property emerges spontaneously from distributional semantics.

A property of certain embeddings that has the potential to help with the above concerns (as well as others) is that of “compositionality”. 

To address the question of compositional structure's presence, we must first look to linguistics and philosophy firstly \citep{andreas2019measuring}. Historically, evaluations of compositionality focused on formal and natural languages \citep{carnap2002logical},\citep{lewis1976general}. These methods, rooted in linguistic representation details like grammar algebra \citep{montague1970universal}, are challenging to apply broadly, especially in non-string-valued spaces.

In the domain of machine learning, the gap in understanding compositionality has elicited a range of scholarly responses. One salient approach is Disentangled Representation Learning (DRL) \citep{bengio2013representation}, conceptualized to discern and segregate intrinsic attributes obfuscated within the representations of manifest data. Such disentangled representations, which can be deconstructed into componential elements, enhance the explicability of the models trained. Each constituent in the latent space pertains to a discrete attribute or feature, thereby streamlining the manipulation and control over data representations. 

\citet{shwartz2019still} undertook an examination of word representation compositionality via six tasks, probing into the phenomena of semantic drift and implicit meaning. \citet{andreas2019measuring} postulated a metric for compositionality based on the approximation fidelity of observed representations when assembled from inferred primitives. This scholar also introduced the Tree Reconstruction Error (TRE) method, focused on gauging the compositionality through multiplication. Notwithstanding these advancements, our scholarly interest predominantly lies in the potential for capturing compositionality within learned data embeddings in their additive manifestation as follows.

A learned representation is compositional when it can represent complex concepts or items by combining simple attributes \citep{fodor2002compositionality}. In this paper, we mainly look into additive compositionality as follows.
$$\textbf{u}_{I} = \sum_{i =1}^{N}\textbf{x}_{i}$$
Where $I$ is an item that has a set of $N$ attributes. $I$ can be represented with embedding vector $\textbf{u}_{I}$, and the attributes can be represented with $\textbf{x}$.

\subsubsection{Compositionality in Sentence Embedding}
Researchers have found that while BERT does not have explicit syntactic trees during training, the representations it learns capture significant syntactic information \citep{hewitt2019structural}. There is an increasing amount of research focusing on evaluating the compositionality in sentence embedding. There are two main approaches: task-based and task-independent. Task-based methods measure the compositionality by evaluating the performance through specific language features, such as semantics, synonym, and polarity. The performance on these tasks defined the compositionality of sentence embedding.

\citet{ettinger2016probing} developed a dataset to identify semantic roles in embeddings, such as whether "professor" is the agent of "recommend". They also looked at semantic scope by altering sentence meanings without much lexical change. \citet{dasgupta2018evaluating} created a dataset examining word combinations in embeddings. They modified sentences to study natural language inference relations, involving changes like word order and addition of words like "more/less" or "not".

These methodologies aim to uncover sentence representation's understanding of language. Task-independent methods, on the other hand, focus on general aspects like sentence length, content, and order.

Without needing specific labeled data, \citet{adi2016fine} presented three evaluation techniques for sentence embeddings: measuring sentence length, identifying a word in a sentence, and determining word order. In tests, LSTM auto-encoders performed well in the latter two tasks.

Nevertheless, no existing research attempts to break down sentence embedding into its attributes. Although \citet{adi2016fine} tried to identify if the building blocks of a sentence, words, were captured by the sentence embedding, the method they used to obtain the word representation was the same as the sentence embedding. Besides, the relation between
these word representations and their corresponding sentence embedding remains unknown.

As a result, in our study, we intend to decompose sentence embedding into word representations and understand if words are the attributes for sentence embedding. Furthermore, the word representation learned from the existing sentence representations can deduce a new sentence embedding. We measure the compositionality by the vector space distance between the actual sentence embedding and the inferred vector that builds from the property vectors.

\subsubsection{Compositionality in Graph Embedding and Algorithm Bias}
The possibility of bias in AI agents has become one of the most significant problems in machine learning. One of the possible sources of bias is the way data is encoded within the agent, and in this paper we are concerned with the possibility that data embeddings contain unwanted information that can lead to what is known as "algorithmic bias". 

As mentioned previously in section \ref{sec: introduction}, we can learn a word's semantic content from the distribution of word frequencies in its context. However, it has been observed that these distributions contain also information of different nature, including associations and biases that reflect customs and practices. For example it is known that the embeddings of color names extracted in this are not gender neutral, nor are those of job titles or academic disciplines. For example, engineering disciplines and leadership jobs may tend to be represented in a "more male" way than artistic disciplines or service jobs \citep{jonauskaite2021english,sutton2018biased, caliskan2017semantics}. 

This could lead to problems that might be described as the machine equivalent of an "unconscious bias", and eventually to unwanted consequences, for example when filtering applicants for a job. 

The presence of gender information in word embeddings was already reported in \citet{bolukbasi2016man}, in an article aptly entitled ``Man is to Computer Programmer as Woman is to Homemaker?''. The same signal was already reported in \citet{mikolov2013distributed}, which introduced the example involving king and queen that we have used above. All this highlights the possibility that  "compositionality" might lead to new ways of reasoning with embeddings, for example by performing analogies. 

An interesting possibility is the presence of similar biases in Knowledge Graph embedding, which would lead both to opportunities and challenges, and which would require attention \citet{Guo2023}. Recent work such as \citet{fisher-etal-2020-debiasing} \citet{bose2019compositional} use adversarial loss to train the model neutral to sensitive attributes. Such a bias can also be observed in movie recommender systems whose embedding is simply trained from a set of movie ratings. Our work discusses new ways to detect it.

\section{Compositionality Detection Methods}
\label{sec: method}
An important consideration is that there is a difference between which information is \textit{present} in a given data representation, and which information is \textit{accessible} to a specific class of functions. While it may be difficult or impossible to prove that certain information is not present, it may be simple to prove that it is not accessible - say  - to a linear function. In practical applications this may be all that is needed. For example, the study \cite{jia2018right} describes a method to ensure that a deep neural network does not contain unwanted information in a form that it can be used by its final - decision making - layers. 

The general problem is as follows. Given a knowledge graph $G = (V, E)$, it may be the case that vertices $V$ have attributes that may be considered private information. For example, suppose we have a graph representing jobs and applicants. Suppose we have vertices representing applicants, vertices representing skills, and vertices representing jobs, with edges denoting which jobs applicants are finally offered. Some attributes of the applicants, for example their gender or age, may be considered \emph{private information} that we do not wish to be able to elicit from the graph.

We give two methods: Correlation-based Fusion Detection and Additive Fusion Detection to detect the fusion of signals in the vertices $V$. We take movie recommender system as a small running example.

\subsection{Correlation-based Fusion Detection}

Canonical Correlation Analysis (CCA) is used to measure the correlation information between two multivariate random variables \citep{shawe2004kernel}. Just like the univariate correlation coefficient, it is estimated on the basis of two aligned samples of observations.

A matrix of binary-valued attribute embeddings, denoted as $\mathbf{A}$, is essentially a matrix representation where each row corresponds to a specific attribute and each column corresponds to an individual data point (such as a word, image, or user). The entries of the matrix can take only two values, typically 0 or 1, signifying the absence or presence of a particular attribute. For example, in the context of textual data, an attribute might represent whether a word is a noun or not, and the matrix would be populated with 1s (presence) and 0s (absence) accordingly.

On the other hand, a matrix of user embeddings, denoted as $\mathbf{U}$, is a matrix where each row represents an individual user, and each column represents a certain feature or dimension of the embedding space. These embeddings are continuous-valued vectors that capture the movie preference of the users. The values in this matrix are not constrained to binary values and can span a continuous range.

Assuming we have a vector for an individual attribute embedding, denoted as 
\[ \mathbf{a} = \left( a_1, a_2, \dots, a_n \right)^T \]
and a vector for an individual user embedding, 
\[ \mathbf{u} = \left( u_1, u_2, \dots, u_m \right)^T \]
our goal is to explore the correlation between these two vectors. To achieve this, we focus on finding projection vectors, \(\mathbf{w}_a\) (where \(\mathbf{w}_{a_k} \in \mathbb{R}^n\)) for the attribute and \(\mathbf{w}_u\) (where \(\mathbf{w}_{u_k} \in \mathbb{R}^m\)) for the user, such that the correlation between the transformed embeddings is maximized. Mathematically, this can be expressed as:
\begin{equation}
    \rho = \max _{\left(\mathbf{w}_{a_k}, \mathbf{w}_{u_k}\right)} \operatorname{corr}\left( \mathbf{w}_{a_k}^T \mathbf{a} , \mathbf{w}_{u_k}^T \mathbf{u} \right)
    \label{eqn: correlation1}
\end{equation}

Note there are $k$ correlations corresponding to $k$ components.

By extending the individual user case to all \( q \) users, we can compute the canonical correlations for the entire user base, which provides insights into the relationship between the attribute embeddings and user embeddings across the whole dataset.

Given two matrices, one representing binary-valued attribute embeddings and the other representing user embeddings, we aim to find a correlation between them. Specifically, we define:

\begin{itemize}
    \item \( \mathbf{A} \): An \( n \times q \) matrix of binary-valued attribute embeddings, where each column represents the attribute embeddings for a specific user, and \( n \) is the number of attributes.
    \item \( \mathbf{U} \): An \( m \times q \) matrix of user embeddings, where each column represents the embedding of a different user, and \( m \) is the dimensionality of each user embedding.
\end{itemize}

To compute the correlation between these matrices, we seek projection matrices \( \mathbf{W}_A \) and \( \mathbf{W}_U \) that maximize the correlation between the transformed \( \mathbf{A} \) and \( \mathbf{U} \). Formally, the objective is:

\begin{equation}
    \rho =\max _{\left(\mathbf{W}_{A}, \mathbf{W}_{U}\right)} \operatorname{corr}\left( \mathbf{A} \mathbf{W}_A, \mathbf{U} \mathbf{W}_U\right)
    \label{eqn:correlation_matrix}
\end{equation}

These paired random variables are often different descriptions of the same object, for example genetic and clinical information about a set of patients \citep{seoane2014canonical}, french and English translations of the same document \citep{NIPS2002_d5e2fbef}, and even two images of the same object from different angles \citep{guo2019canonical}.

In the example of viewers and movies, we use this method to compare two descriptions of users. One matrix is based on demographic information, which are indicated by Boolean vectors. The other matrix is based on their behaviour, which is computed by their movie ratings only.

\begin{figure}[htbp]
        \centering
        \includegraphics[scale=.20]{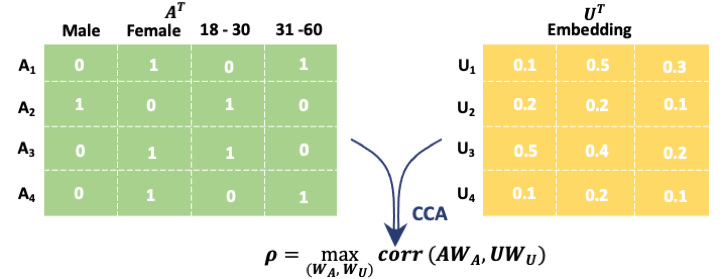}
         \caption{Schematic of Correlation-based Fusion Detection}
          \label{fig: CCA}
\end{figure}

\subsection{Additive Fusion Detection}
\label{sec: linearsys}
Again assuming we have a matrix of entity embeddings $\mathbf{U}$ with matrix of attributes $\mathbf{A}$, we investigate the possibility that the entity embeddings can be decomposed into a linear combination of embeddings corresponding to attributes. Specifically, we investigate whether we can learn a matrix $\mathbf{X}$ as follows
\begin{equation}
  \mathbf{A}\mathbf{X} = \mathbf{U}
\end{equation}

As mentioned in Section \ref{sec: section2}, word embeddings generated from the distribution of words in text can encode additional semantic or syntactic information.
We investigate here the possibility that entity embeddings in knowledge graphs can be decomposed into linear combinations of embeddings corresponding to attributes. We use methods from \citet{xu2023on} to see if an entity embedding $\mathbf{u}$ can be decomposed into a linear system.

In our example of viewers and movies, a set of users as $U$ and the coefficient matrix of the components as $\mathbf{A}$. We aim to solve a linear system $\mathbf{A\mathbf{X}}=\mathbf{U}$ so that the user embedding can be decomposed into three components (gender, age, occupation) as follows, $\mathbf{u} = \sum_{i} a_i \mathbf{x}_i$. Here, $\mathbf{u}$ is a user embedding, $i$ ranges over all possible values of each private attribute, $\mathbf{x}_i $ is an embedding corresponding to the $i$th attribute value, and $a_i\in\{0, 1\}$, denotes whether a particular attribute value is present or absent for the user. This formulation allows us to break down each user into distinct, quantifiable components, reflecting their demographics and interests.
\begin{figure}[htbp]
         \centering\includegraphics[scale=.15]{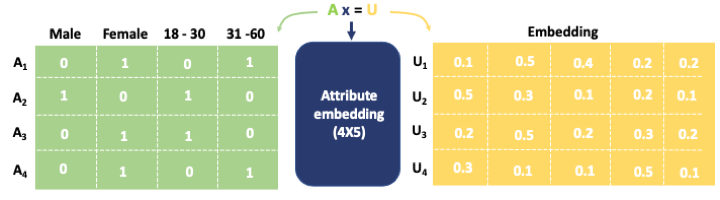}
         \caption{Schematic of Additive Fusion Detection: our linear decomposition system}
          \label{fig: linear}
\end{figure}

\subsection{Hypothesis Testing with Random Permutations}
\subsubsection{Methods}
\label{sec:stattest}
We aim to investigate the correlation between user attributes and their movie preferences. By measuring a test statistic for correlation, and subsequently employing a permutation test on one of the datasets, we assess the likelihood of observing the same degree of correlation under the null hypothesis of no association.

To assess the significance of the observed correlation, a permutation test was conducted. This involved randomizing the order of users in one of the datasets (either attributes or movie preferences) while keeping the order in the other dataset unchanged. The test statistic for correlation was recalculated for each permutation.

Our null hypothesis is that the embedding of a vertex $u$ and its attributes $a$ are independent. To test whether this is the case, we employ a non-parametric statistical test, whereby we directly estimate the $p$-value as the probability that we could obtain a ``good''\footnote{either high or low, depending on the statistic} value of the test statistic under the null hypothesis. If the probability of obtaining the observed value  of the test statistic is less that 1\%, we reject the null hypothesis. 

Specifically, we will randomly shuffle the pairing of vertices and attributes 100 times, and compute the same test statistic. If the test statistic of the paired data is better than that of the randomly shuffled data across all 100 random permutations, we conclude that the correctly paired data performs better to a 1\% significance level.

The test statistic for \emph{Correlation-based Fusion Detecion} is the correlation $\rho$ For the \emph{Additive Fusion Detection} $\mathbf{A}\mathbf{X}=\mathbf{U}$, we use the Leave One Out algorithm as shown in Algorithm \ref{alg: LOO}, that is to leave one user out and predict either the user embedding or the inverse problem of user identity. We look at the L2 norm loss of the linear system, cosine similarity and retrieval accuracy, a metric defined in \citet{xu2023on}.

\begin{itemize}
        \item L2 Loss of the linear system $||\mathbf{A}\mathbf{X} - \mathbf{U}||^2$
        \item Cosine similarity between $\mathbf{u}$ and constructed embedding $\hat{\mathbf{u}}$
        \item Accuracy of retrieving identity of $\mathbf{u}$ with $\hat{\mathbf{u}}$
    \end{itemize}

\begin{figure}[htbp]
         \centering
         \includegraphics[scale=.30]{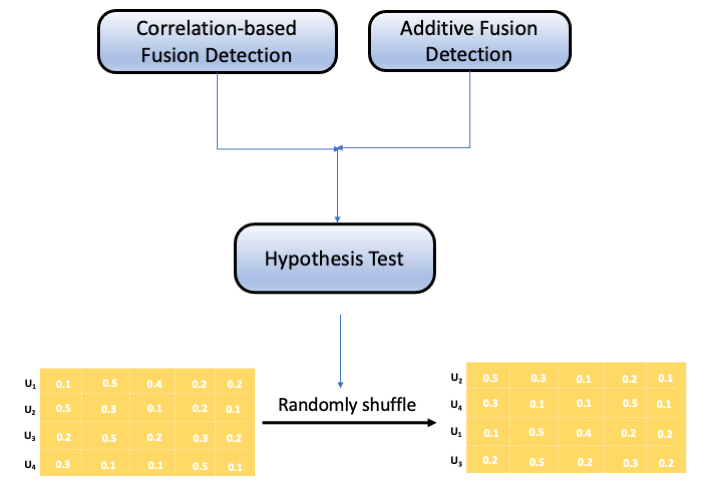}
         \caption{Hypothesis Testing}
          \label{fig: Hypothesis}
\end{figure}

\begin{algorithm}

\caption{Leave One Out}
\label{alg: LOO}
\begin{algorithmic}[1]
\For{any dataset of ($\mathbf{A}, \mathbf{U}$) descriptions} \Comment{(*)}
    \For{each user \( u \)}
        \State Leave the user \( u \) out
        \State Train on the remaining \( N-1 \) users
        \State Predict the user behavior $\hat{\mathbf{U}}$ \Comment{(**)indicate the synthetic/predicted behavior with $\hat{}$}
        \State Measure the quality of $\hat{\mathbf{U}}$ \Comment{(***)}
    \EndFor
    \State The Score is average quality (across all users) of artificial embeddings $\hat{\mathbf{U}}$
\EndFor
\end{algorithmic}
\end{algorithm}

\begin{algorithm}
\caption{Compute a Loss Function}
\label{alg:loss}
\begin{algorithmic}[1]
\State For a specific user, with true behavior \(\mathbf{U} \) and predicted behavior $\hat{\mathbf{U}}$
\begin{itemize}
    \item L2 Norm between \(\mathbf{U} \) and $\hat{\mathbf{U}}:||\mathbf{A}\mathbf{X} - \mathbf{U}||^2$
    \item Cosine between \(\mathbf{U} \) and $\hat{\mathbf{U}}$
    \item Identity between \(\mathbf{U} \) and best\_match\_of: $\hat{\mathbf{U}}$
\end{itemize}
\end{algorithmic}
\end{algorithm}

\noindent \textbf{Notes:} \\
(*) This includes randomly shuffled ($\mathbf{A}, \mathbf{U}$) pairs. \\
(**) Here, we take use as an example, the user behavior means user embedding computed by the movie preference, it could also be word/sentence embedding computed by the context. \\
(***) This includes different loss functions as shown in Alogorithm \ref{alg:loss}.

\subsubsection{Analysis}
\paragraph{Hypothesis testing on Correlation-based Fusion Detection}
In this study, we employ a non-parametric testing approach to directly estimate the p-value as the probability of an event under the null hypothesis. This event pertains to the chance occurrence of a high value of the test statistic, specifically a strong correlation between two datasets. By leveraging a Monte Carlo sampling method, where random permutations of the user list serve as the basis for our samples, we assess the likelihood of observing the given test statistic purely by chance. If the probability of achieving the observed test statistic is less than 1\%, we lean towards rejecting the null hypothesis. However, it is important to note that this does not conclusively affirm the alternative hypothesis ($H_{1}$) but rather emphasizes the statistical significance of our findings, a nuance that delves into the philosophical underpinnings of statistical inference.
\paragraph{Hypothesis testing on Additive Fusion Detection}
In this segment of the study, our objective is to substantiate the hypothesis that the embedding of user behavior can be characterized by user demographics. We postulate that the representation of user behavior, termed here as the "user-behavior-embedding", can be approximated as a summation of vectors representing user demographics. To evaluate the accuracy of this approximation, we employ a test statistic based on the loss or distance between the actual user behavior embedding and its demographic-based approximation. A critical inquiry that emerges is: given the computed loss value, what is the probability that such a value could arise purely by chance under the null hypothesis? To address this, we implement a permutation-based approach, wherein we shuffle the data and estimate the probability of obtaining our observed test statistic under randomized conditions.

\section{Experimental Study }
\ml{give a short intro paragraph describing the studies we will look at}
We will examine the semantic and syntactic signals in word2vec embeddings, comparing them to WordNet and MorphoLex benchmarks. Subsequently, we will analyze the compositionality of BERT sentence embeddings, hypothesizing an additive relationship between individual word and complete sentence representations. Finally, using the MovieLens dataset, we will study the relationship between user movie preferences and demographic traits through behavior-based embeddings.
\label{sec: experiments}

\subsection{Word embedding}
In our investigation, we will be particularly interested in examining two distinct signals encapsulated within the word2vec embeddings: semantic and syntactic information. To discern these signals, we employ WordNet embeddings as a benchmark for semantic representation, while MorphoLex serves as our reference for syntactic structures. By comparing the word2vec embeddings against both WordNet and MorphoLex, we are able to disentangle and analyze the semantic and syntactic nuances inherent in the word2vec representation. This comparative approach provides a comprehensive understanding of the multifaceted linguistic properties embedded within word2vec.

\subsubsection{WordNet}
WordNet \citep{miller1995wordnet} is a large lexical database of English, which consists of 40943 entities and 11 relations. Synsets are interlinked by means of conceptual-semantic and lexical relations. WordNet is a combination of dictionary/thesaurus with a graph structure. Nouns, verbs, adjectives, and adverbs are grouped into sets of cognitive synonyms (synsets), each expressing a distinct concept. These synsets are interlinked using conceptual-semantic and lexical relations.

The relations include, for instance, synonyms, antonyms, hypernyms (kind of relationship), hyponyms (part of relationship), meronyms (member of relationship), and more. For example, searching for `ship' in WordNet might yield relationships to `boat' (as a synonym), `cruise' (as a verb related to `ship'), or `water' (as a related concept), among other things.

\paragraph{Mapping Freebase ID to text}
WordNet is constructed with Freebase ID only, an example triple could be <00260881, hypernym, 00260622>. We follow \citet{villmow2019} to preprocess the data and map each entity with the text with a real meaning. 

The above triple can then be processed with the real semantic meaning: <land reform, hypernym, reform>. The Word2Vec word embedding is pretrained from a google news corpus. 
\subsubsection{WordNet Embedding}
We want to ensure our WordNet embedding can contain the semantic relation in it. Therefore, we train the embedding with the task of predicting the tail entity given a head entity and relation. For example, we want to predict the hypernym of piciform bird:
$$< \textbf{piciform bird}, hypernym, \textbf{?}>$$
We train the WordNet Embedding in the following way:
\begin{enumerate}
    \item We split our dataset to use 90\% for training, 10\% for testing.
    \item Triples of $\left(head, relation, tail\right)$ are encoded as relational triples $\left(h, r, t\right)$.
    \item We randomly initialize embeddings for each $h_i$, $r_j$, $ t_k$ and use the scoring function in Equation \ref{transe} and minimize the loss by Margin Loss.
    \item We sampled 20 corrupted entities. Learning rate is set at 0.05 and training epoch at 300. 
\end{enumerate}
Detailed results can be found in the Table \ref{tab: wordnet}, which shows that our WordNet embeddings do contains the semantic information.

\begin{table}[width=.9\linewidth,cols=4,pos=h]
\centering
\caption{link prediction performance for WordNet}
\begin{tabular}{@{}lllcc@{}}
\toprule
 & Hits@1 & Hits@3 & Hits@10 & MRR\\ \midrule
WordNet & 0.39 & 0.41 & 0.43 & 0.40 \\  \bottomrule
\end{tabular}
\label{tab: wordnet}
\end{table}
\subsubsection{MorphoLex}
MorphoLex\citep{sanchez2018morpholex} provides a standardized morphological database derived from the English Lexicon Project, encompassing 68,624 words with nine novel variables for roots and affixes. Through regression analysis on 4724 complex nouns, the dataset highlights the influence of root frequency, suffix length, and the prevalence of frequent words in a suffix's morphological family on lexical decision latencies. It offers valuable insights into morphology's role in visual word processing.

In this paper, we specifically focus on words with one root and multiple suffixes. For the CCA experiment, words with suffixes occurring less than 10 times are filtered out. Conversely, in the linear decomposition experiment, we exclude rows with roots appearing fewer than 3 times. 

\begin{table}[]
    \centering
        \caption{Suffix presence (indicated by '1') for selected words from the MorphoLex dataset, see \url{https://github.com/ZhijinGuo/Compositional-Fusion-of-Signals-in-Data-Embedding} for full table}
    \begin{tabular}{lrrrrrr}
    \toprule
               Word &  al &  ic &  ist &  ity &  ly &  y \\
    \midrule
      allegorically &   1 &   1 &    0 &    0 &   1 &  0 \\
     whimsicalities &   1 &   0 &    0 &    1 &   0 &  1 \\
       whimsicality &   1 &   0 &    0 &    1 &   0 &  1 \\
        whimsically &   1 &   0 &    0 &    0 &   1 &  1 \\
    voyeuristically &   0 &   1 &    1 &    0 &   1 &  0 \\
    \bottomrule
    \end{tabular}

    \label{tab:MorphLex}
\end{table}

\subsubsection{Correlation-based Fusion of Semantic and Morphology in Word2Vec}
\begin{figure}[htbp]
        \centering
         \includegraphics[scale=.50]{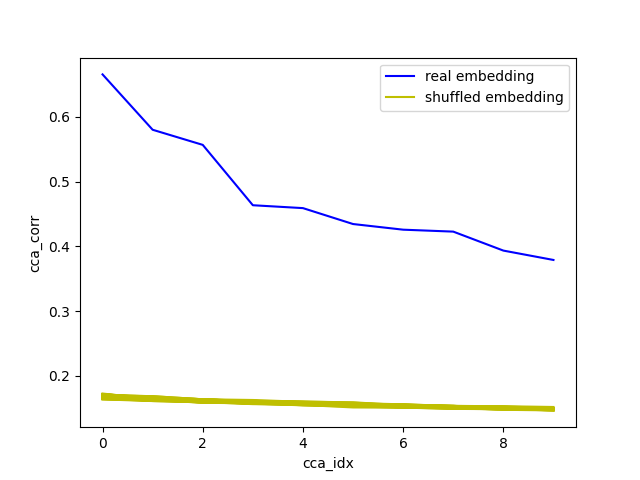}
         \caption{PCC for the true WordNet-Word2Vec pairings and 100 permuted pairings, the first 10 components are selected for illustration. PCC is calculated between projected $\mathbf{A}$ and projected $\mathbf{U}$. $x$ axes stands for the $k$th components, $y$ axes gives the value. The PCC value for real pairings is larger than for any permuted pairings.}
         \label{fig: PCC_wordnet}
         \end{figure}

We applied Correlation-based Fusion Detection to compare two different representations of a set of words. Word2Vec provides a vector space model that represents words in a high-dimensional space, using the context in which words appear. 
\paragraph{Semantic}
WordNet offers a structured lexical and semantic resource where words are related based on their meanings and are organized into synonym sets. We shuffled the pairing of Word2Vec embedding and words 100 times to break the semantic signal captured in the Word2vec embedding, the result is shown in Figure \ref{fig: PCC_wordnet}.

\begin{figure}[htbp]
        \centering
         \includegraphics[scale=.50]{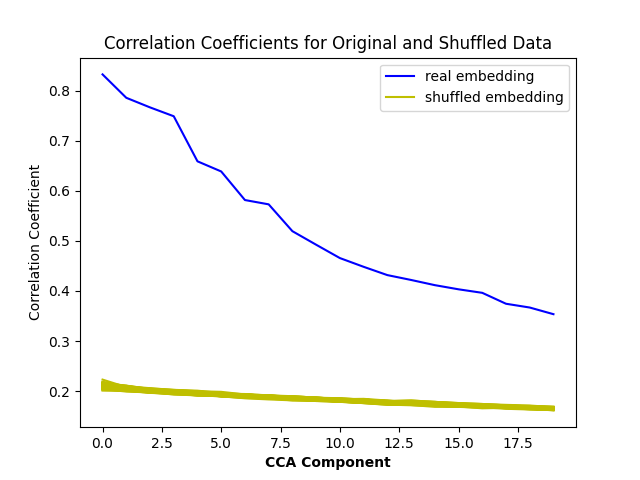}
         \caption{PCC. comparasion for the true MorphoLex-Word2Vec pairings and 100 permuted pairings, the first 20 components are selected for illustration. PCC is calculated between projected $\mathbf{A}$ and projected $\mathbf{U}$. $x$ axes stands for the $k$th components, $y$ axes gives the value. The PCC value for real pairings is larger than for any permuted pairings.}
         \label{fig: PCC_morphology}
         \end{figure}

The correlation between two different representations is higher than the shuffled ones in the first component, which means, the structured semantic information can be captured from the word embedding trained by its context words.
\paragraph{Morphology}
Conversely, MorphoLEx provides a morphological resource predicated on root frequency, suffix length, and the function of morphology. For experimental robustness, we permuted the Word2Vec embedding on 50 separate occasions to obfuscate the morphological signals intrinsic to the Word2Vec representation, with results delineated in Figure \ref{fig: PCC_morphology}.

The correlation coefficient observed between the two distinct representations surpasses that of the permuted counterparts in the principal component. This suggests that morphological nuances are ascertainable from word embeddings informed by their contextual counterparts.

\subsubsection{Decomposing Word2Vec Embedding by Addtive Fusion Detection}
We have chosen a collection of 278 words, where several words have common roots, and others have identical morphological units. Having computed a set \(\mathbf{U} \in \mathbb{R}^{278 \times 300}\) of embeddings as Word2Vec embeddings, we can find the unknown vectors \(\mathbf{x}_{i}\), \(\mathbf{x}_{j}\), and \(\mathbf{x}_{k}\) by solving the linear system \(\mathbf{AX} = \mathbf{U}\), where \(\mathbf{A} \in \mathbb{R}^{278 \times 45}\) is a binary matrix indicating the presence or absence of each root words and morphemes, This system does not have (in general) an exact solution, so we approximate the solution by solving a linear least squares problem, using the pseudo-inverse method, as follows:
\begin{equation}
\label{eqn:x}
\textbf{X} = (\textbf{A}^{T}\cdot \textbf{A})^{-1}\cdot \textbf{A}^{T}\cdot \textbf{U} 
\end{equation}

In our leave-one-out approach, we train the linear system without including the target word $u$, allowing us to generate root words and morphemes independently of $u$. We test the accuracy of this method by estimating the embedding for a new word and comparing it to its true Word2Vec embedding, using the evaluation steps outlined in Algorithm \ref{alg:loss}.

\begin{figure*}[!h]
    \centering
    \begin{subfigure}[H]{0.32\textwidth}
        \centering
        \includegraphics[width=\textwidth]{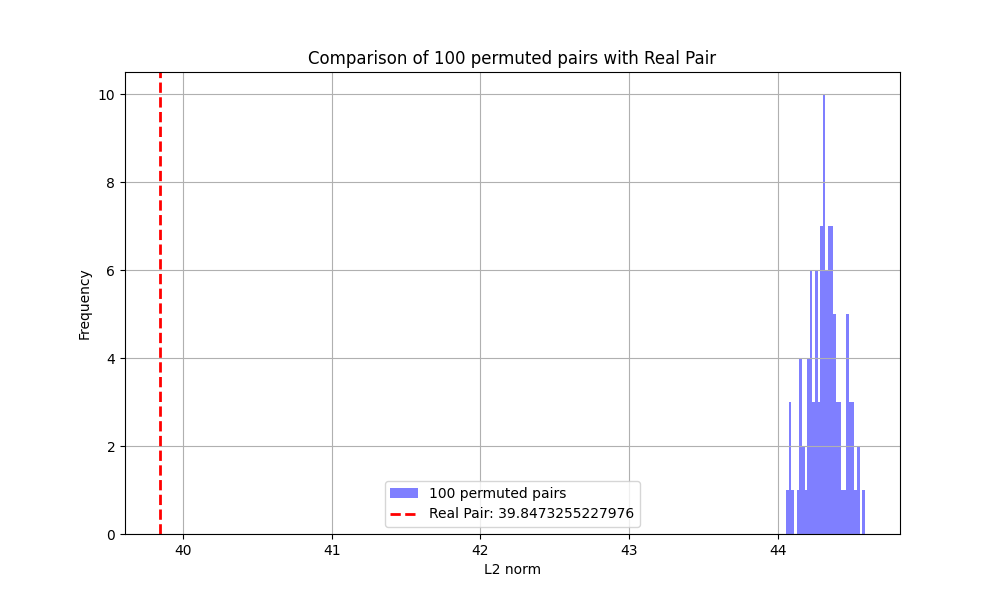}
        \centering
        \caption{Linear System Loss}
    \end{subfigure}
    \begin{subfigure}[H]{0.32\textwidth}
        \centering
        \includegraphics[width=\textwidth]{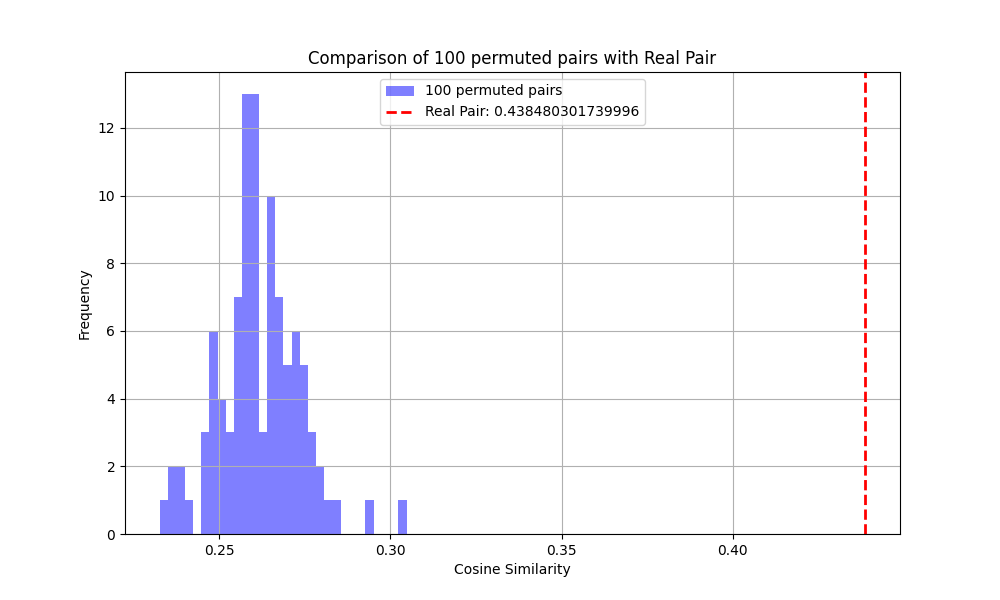}
        \caption{Cosine Similarity}
    \end{subfigure}
    \begin{subfigure}[H]{0.32\textwidth}
        \centering
        \includegraphics[width=\textwidth]{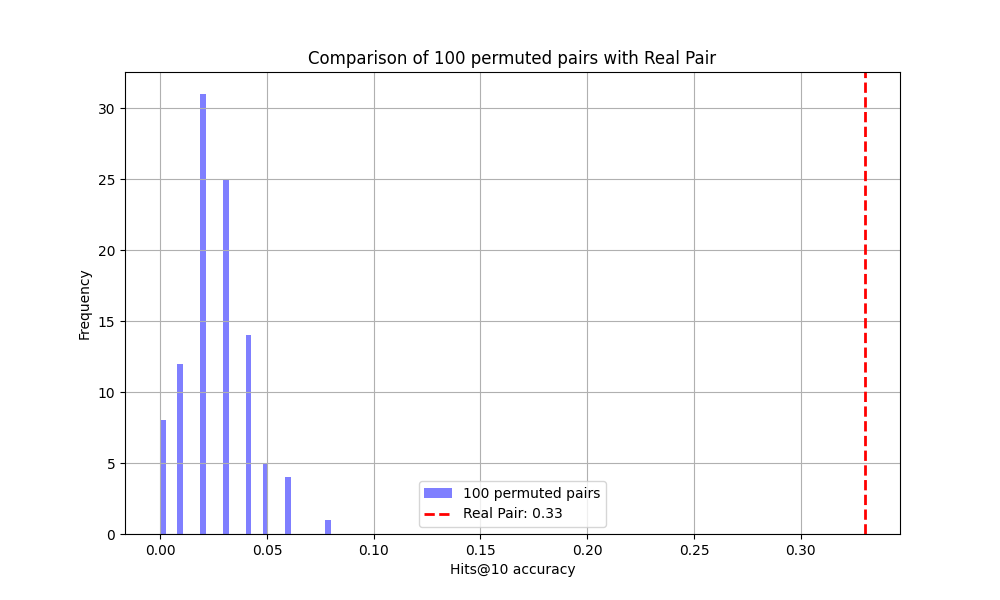}
        \caption{Retrieval Accuracy@1}
    \end{subfigure}
    \caption{The test statistics for Word2vec embedding decomposition. Dash line is the average performance of $ \hat{\textbf{B}} $ learned from the Word2Vec embedding. The bars are the distribution of the results from random permutations that run for 100 times. }
    \label{fig:word2vec_results}
\end{figure*}

Figure \ref{fig:word2vec_results} delineates the efficacy of decomposing the Word2vec embedding. The results show that the Word2Vec embedding can be bifurcated into distinct components: the root and the morphemes. These components can subsequently be employed to predict the embedding of novel words.

When the linear system decomposes the Word2Vec embedding, it incurs a loss of 38.85. Notably, this is more efficient than the minimum loss observed from random permutations, which stands at 44.06. Consequently, the p-value from non-parametric testing falls below the significance threshold ($\alpha$=0.01), leading us to reject $H_{0}$. This suggests that the Word2Vec embedding can be conceptualized as an amalgamation of two discrete attributes.

Furthermore, it's feasible to approximate the embedding of a word using solely the root and morphological suffix components derived from the linear system. Such a reconstructed embedding, denoted as $\hat{U}$, can be compared to reconstructions based on randomized (attributes, embeddings) pairs using cosine similarity as the metric. Intriguingly, the cosine similarity between the authentic embedding and $\hat{U}$ is 44\%, surpassing all instances from random permutations.

The efficacy of the reconstructed embedding is further underscored by its ability to retrieve the actual embedding with a hits@10 accuracy of 33\%. In stark contrast, embeddings composed with randomized attribute/embedding pairs demonstrate a paltry retrieval success, peaking at a mere 8

\subsection{Sentence Embedding}
\label{sec:svo}
Following the decompostion for Word2Vec embeddings,  we have further interests if sentence embedding can be decomposed in a similar way. Sentences are compositional structures that are built from words. Therefore, it is natural to ask if the learned representations reflect the compositionality. We assume that there is an additive compositionality between words and sentences so that the sentence representation can be decomposed in terms of $$\Phi_{BERT}(Sentence) \approx \Phi(Word_{1}) + \dots + \Phi(Word_{N})$$

We leverage a linear system to decompose the sentence embedding into word representations to investigate the compositionality in BERT sentence embedding. To do this, we generated a sentence corpus that includes 1,000 sentences. Each sentence consists of the simplest elements required for completing a sentence: subject, verb and object.

\subsubsection{Data Generation}
We constructed a sentence corpus\footnote[1]{The corpus can be accessed at \url{https://github.com/CarinaXZZ/On_Compositionality_in_Data_Embedding}} with 30 distinct components categorized into subjects (Sbj), verbs, and objects (Obj), which we then arranged into 10x10x10 triplet combinations of $(Sbj, Verb, Obj)$. These triplets form short sentences utilizing consistent prepositions and articles. For instance, the triplet $(cat, sat, mat)$ yields the sentence ``The cat sat on the mat.'' Our corpus comprises 1000 such sentences, enabling detailed analysis of each component's role when decomposing with a linear system.

BERT employs a subword tokenization strategy, splitting words like ``bookshelf'' into ``book'' and ``shelf''. We selected corpus words to maintain uniform token counts across sentences. Since BERT considers punctuation as tokens, each sentence amounts to seven tokens.

To construct a sentence (I), we add the subject, verb, and object phrases with indices i, j, and k, respectively. Thus, $I_{i,j,k} = Sbj_{i} + Verb_{j} + Obj_{k}$. We calculate sentence embedding $\textbf{U}_{i, j, k} = \Phi_{BERT} (I_{i,j,k})$ with a fine-tuned BERT introduced in section \ref{sec:se}.

\subsubsection{Decomposing Sentence BERT Embedding by Additive Fusion Detection}

Given a set of sentence embeddings \(\mathbf{U}\), we determine the unknown vectors \(\mathbf{x}_{i}\), \(\mathbf{x}_{j}\), and \(\mathbf{x}_{k}\) by resolving \(\mathbf{AX} = \mathbf{U}\). Here, \(\mathbf{A}\) is a \(1000 \times 30\) binary matrix specifying each sentence component, \(\mathbf{X}\) represents the \(30 \times 768\) BERT embeddings for sentence attributes, and \(\mathbf{U}\) is the \(1000 \times 768\) matrix of sentence embeddings. The solution is obtained via the pseudo-inverse method, The embedding accuracy is quantified by the loss \( L \), defined as:
\begin{equation}
    \label{eqn:norm}
    L = \| \mathbf{AX} - \mathbf{U} \|^{2}
\end{equation}
For our null hypothesis, sentence embeddings are randomized to disrupt the sentence-embedding association, and loss is computed for this perturbed data over 100 iterations.

One of the interesting challenges is if we can predict the sentence embedding $\textbf{u}$ with the word representations solved by the linear system without seeing the actual sentences. To test this, we utilise the leave one out method to solve the linear system and reconstruct the sentence embedding by adding up the word representations we obtained with equation \ref{eqn:x} so that 
\begin{equation}
\begin{gathered}
    \label{eqn:predict}
    \Phi_{C}(I) =  \Phi_{C}(Sbj) + \Phi_{C}(Verb) + \Phi_{C}(Obj) \\
    \end{gathered}
\end{equation}
Here $\Phi_{C}$ represents the composed embedding $\Phi_{Composed}$.
We again apply the leave-one-out strategy, excluding the target sentence $I$ from the dataset while training the linear system. This approach ensures word representations are formed with no foreknowledge of $I$. The efficacy of these elements is evaluated by predicting a new sentence's embedding, then measuring its likeness to the actual BERT embedding. We assess this through two methods: first, by calculating the cosine similarity between the predicted and real embeddings; second, by determining if the predicted embedding can identify the correct sentence among 1000 possibilities. Each round involves omitting a sentence, solving the linear system with the rest, and then using the deduced components to estimate its embedding.

\subsubsection{Results}

\begin{figure*}[!h]
    \centering
    \begin{subfigure}[H]{0.32\textwidth}
        \centering
        \includegraphics[width=\textwidth]{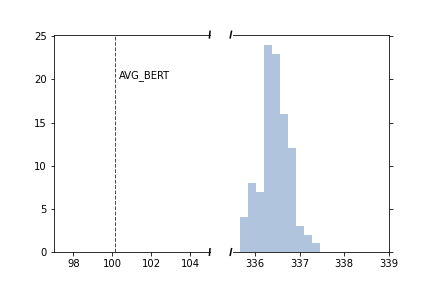}
        \centering
        \caption{Linear System Loss}
    \end{subfigure}
    \begin{subfigure}[H]{0.32\textwidth}
        \centering
        \includegraphics[width=\textwidth]{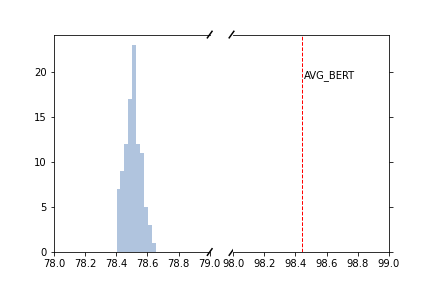}
        \caption{Cosine Similarity}
    \end{subfigure}
    \begin{subfigure}[H]{0.32\textwidth}
        \centering
        \includegraphics[width=\textwidth]{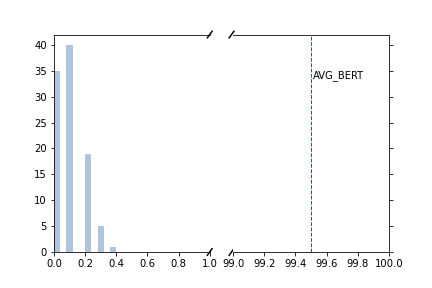}
        \caption{Retrieval Accuracy@1}
    \end{subfigure}
    \caption{The test statistics for sentence embedding decomposition. AVG\_BERT is the average performance of $ \hat{\textbf{B}} $ learned from the BERT embedding. The bars are the distribution of the results from random permutations that run for 100 times \citep{xu2023on}. }
    \label{fig:se_results}
\end{figure*}

Figure \ref{fig:se_results} illustrates the performance of decomposing BERT sentence embedding. These results show that the BERT sentence embedding can be decomposed into three separate components: subject, verb, and object. And those components can then be used to predict the embedding of a new sentence. 

The sentence embedding decomposition via the linear system yields a minimal loss of 100.14, significantly less than the smallest loss from random permutations at 335.65. This results in a p-value below the significance level \(\alpha = 0.01\), leading to the rejection of \(H_{0}\). Consequently, BERT sentence embeddings are effectively representable by the sum of their Sbj, Verb, and Obj components.

The sentence's embedding, denoted as \(\hat{\mathbf{U}}\), can be approximated using the Sbj, Verb, Obj components obtained from the linear system. This approximated embedding \(\hat{\mathbf{U}}\) exhibits a 98.44\% cosine similarity with the BERT embedding, surpassing all comparisons with randomized trials.

Furthermore, \(\hat{\mathbf{U}}\) achieves a 99.5\% success rate in retrieving the correct BERT embedding, whereas the best retrieval accuracy using randomized attribute/embedding pairings does not exceed 0.4\%.

\subsection{Graph Embedding}
Leveraging the MovieLens dataset, we employ graph embeddings to compute user representations based on their movie preferences. Our primary objective is to uncover demographic signals that might be implicitly captured within these behavior-based embeddings. To achieve this, we juxtapose the computed user embeddings against a boolean matrix representing demographic information. By analyzing the correlation between the embeddings and the demographic matrix, we aim to elucidate the extent to which user behavior, as manifested in movie preferences, aligns with or diverges from demographic characteristics. We train our model on GeForce GTX TITAN X.
\subsubsection{Datasets}
\label{sec:movielenstraining}
This experiment was conducted on the MovieLens 1M dataset \citep{harper2015movielens} which consists of a large set of movies and users, and a set of movie ratings for each individual user. It is widely used to create and test recommender systems. Typically, the goal of a recommender system is to predict the rating of an unrated movie for a given user, based on the rest of the data. In particular, there are 6040 users and approximately 3900 movies. Each user-movie rating can take values in 1 to 5, 1 representing a low rating and 5 a high rating. There are 1 million triples (out of a possible $6040\times3900 = 23.6m$), so that the vast majority of user-movie pairs are not rated. 

Users and movies each have additional attributes attached. For example, users have demographic information such as gender, age, or occupation. Whilst this information is typically used to improve the accuracy of recommendations, we use it to test whether the embedding of a user correlates to private attributes, such as gender or age. We therefore compute our graph embedding based only on ratings, leaving user attributes out. Experiments for training knowledge graph embeddings are implemented with the OpenKE \citep{han-etal-2018-openke} toolkit.

We embed the knowledge graph in the following way:
\begin{enumerate}
    \item We split our dataset to use 90\% for training, 10\% for testing.
    \item Triples of $\left( user, rating, movie\right)$ are encoded as relational triples $\left(h, r, t\right)$.
    \item We randomly initialize embeddings for each $h_i$, $r_j$, $ t_k$ and train embeddings to minimize the loss in equation \eqref{eq:actual_loss}.
    \item We sampled 10 corrupted entities and 4 corrupted relations. Learning rate is set at 0.01 and training epoch at 300. 
\end{enumerate}
We verify the quality of the embeddings by carrying out a link prediction task on the remaining 10\% test set. We achieved a RMSE score of 0.88, Hits@1 score of 0.46 and Hits@3 as 0.92, MRR as 0.68 and MR as 1.89.

\begin{figure}[htbp]
  \centering
  \includegraphics[width=\linewidth]{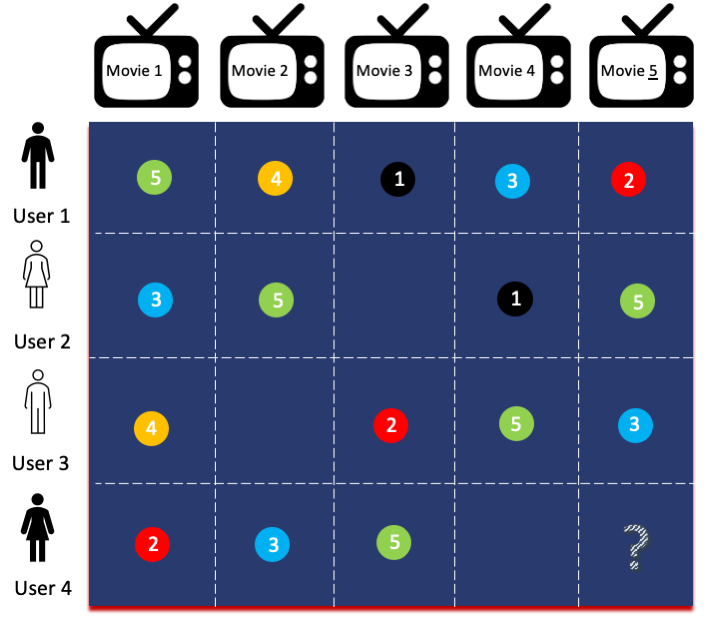}
  \caption{An illustration of a movie rating system}
  \label{fig: movie recommender system}
\end{figure}

We trained our model on 90\% of the available triples and predicted the remaining 10\% missing ones (missing edges or links or relations). We sampled 10 corrupted entities, and 4 corrupted relations, with setting the learning rate as 0.01 and training epoch as 300. 

Recall that we trained embeddings on the MovieLens dataset without including any user information. We now apply our three methods for bias detection to investigate the extent to which private information can be detected. 
\subsubsection{Correlation-based Fusion Detection}
We collect attribute information for all 6040 users and embed their personal attributes with Boolean indicator vectors $\mathbf{a}_i$ which encode the value of each attribute (gender, age, and occupation). 
We investigate whether users' private traits may be leaked from the graph embeddings by comparing two different user representations $\mathbf{a}_i$, the Boolean vector of attributes, and $\mathbf{u}_i$, the user embedding calculated as in section \ref{sec:movielenstraining}. 

\begin{figure}[htbp]
         \centering
         \includegraphics[scale=.40]{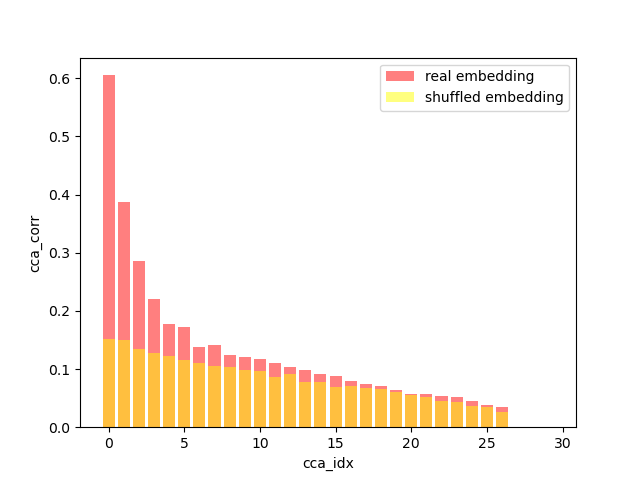}
         \caption{Pearson's correlation coefficient (PCC) for true user-attribute pairings and 100 permuted pairings. PCC is calculated between projected $\mathbf{A}$ and projected $\mathbf{U}$. $x$ axes stands for the $k$th components, $y$ axes gives the value. The PCC value for real pairings is larger than for any permuted pairings.}
          \label{fig: p-value-kg}
\end{figure}

We apply CCA to calculate the correlation between users and their attributes. We apply the non-parametric statistical test described in section \ref{sec:stattest}. Specifically, our null hypothesis is that users' movie preferences are not correlated with their attributes. We calculate Pearson's correlation coefficient (PCC) between projected $\mathbf{A} \mathbf{w}_A$ and projected $\mathbf{U} \mathbf{w}_U$. We go on to calculate the PCC between 100 randomly generated pairings of user and attribute embeddings, and find that the PCC between true pairs of attribute and user embeddings is higher each time. We therefore reject the null hypothesis at a 1\% significance level. 
The correlation coefficients between real pairs and random pairs is reported in figure \ref{fig: p-value-kg}. 

\begin{figure}[htbp]
         \centering
         \includegraphics[scale=.50]{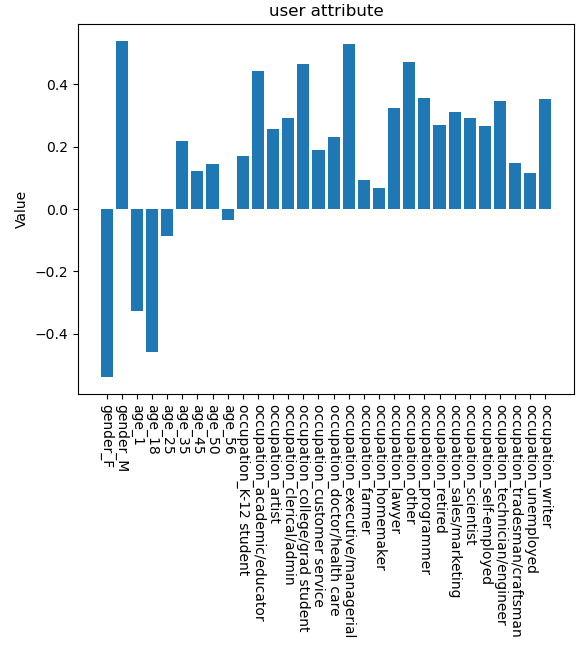}
         \caption{Distribution for each attribute on the second component of CCA}
         \label{fig: genderdistribution}
\end{figure}

Figure \ref{fig: genderdistribution} displays weights indicating the contribution of each component to the overall attribute fusion as determined by the correlation-based fusion detection.

\subsubsection{Additive Fusion Detection on Gender and age}
Preliminary results indicated a certain level of correlation between user attributes and movie preferences as measured by the test statistic. Subsequent permutation tests revealed that the observed correlation was rarely, if ever, achieved under randomized conditions.

We investigate the ability of a user embedding to be reconstructed as a linear sum of attribute embeddings by doing the leave-one-out experiment.
We then try to interpret the knowledge graph embedding with user attributes. Similar to sentence embedding, a linear system is used to calculate the representation for each user attribute. Note that not all of the combinations of attributes
exist in the movie lens dataset. 
We find that a user embedding can be reconstructed as a linear combination of its attributes by solving the linear system described in section \ref{sec: linearsys}. We use the pseudo-inverse method to solve this system.
We try to interpret the user embedding with user attributes such as gender and age.
we first group the user by age and gender firstly and compute the mean embedding of 14 group of users.  We use three test statics as mentioned in Section \ref{sec:stattest} to test our linear system. We set a significance threshold: $\alpha$ = 0.01.

Same as the Correlation-based Fusion Detection setting, we permuted the pairing of users 100 times.
Table \ref{tab: statistic test} shows the observed p- value for three different statistics, which is the probability of seeing that value of statistic under the null hypothesis.
We first decompose the user embedding into gender and age. Our results show the linear system is able to decompose the user embedding with a loss of 0.47 which is lower than every loss for a random permutation (1.11-2.11). The cosine similarity is $99.8\%$, higher than any permuted pairs. The identity retrieval accuracy is 0.79 which is higher than any random permuted pairs (0.0-0.14). Therefore, the null hypothesis is rejected. This shows that a user embedding can be reconstructed as a linear combination of gender and age.

\begin{figure*}[!h]
    \centering
    \begin{subfigure}[H]{0.32\textwidth}
        \centering
        \includegraphics[width=\textwidth]{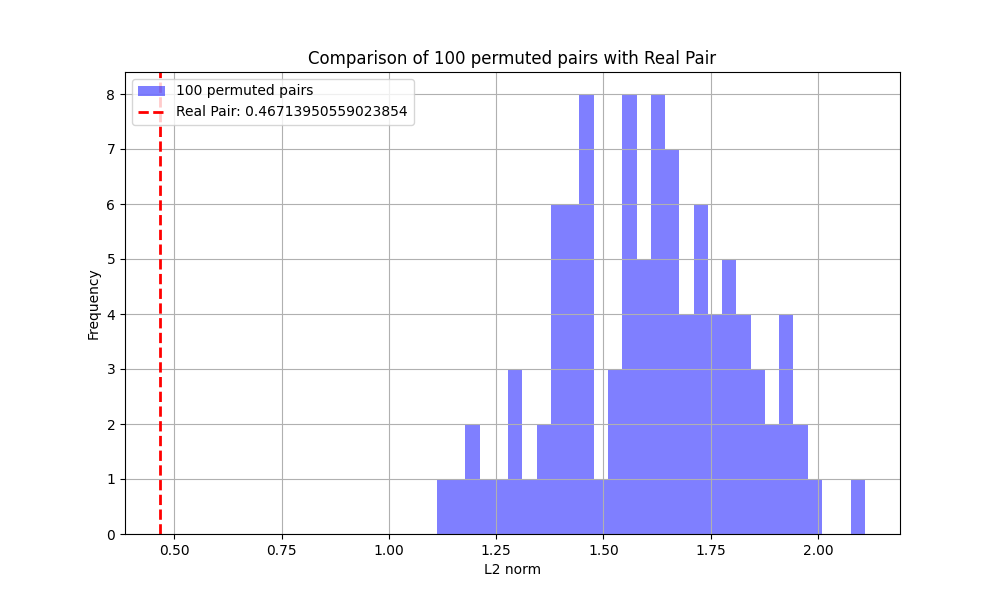}
        \centering
        \caption{Linear System Loss}
    \end{subfigure}
    \begin{subfigure}[H]{0.32\textwidth}
        \centering
        \includegraphics[width=\textwidth]{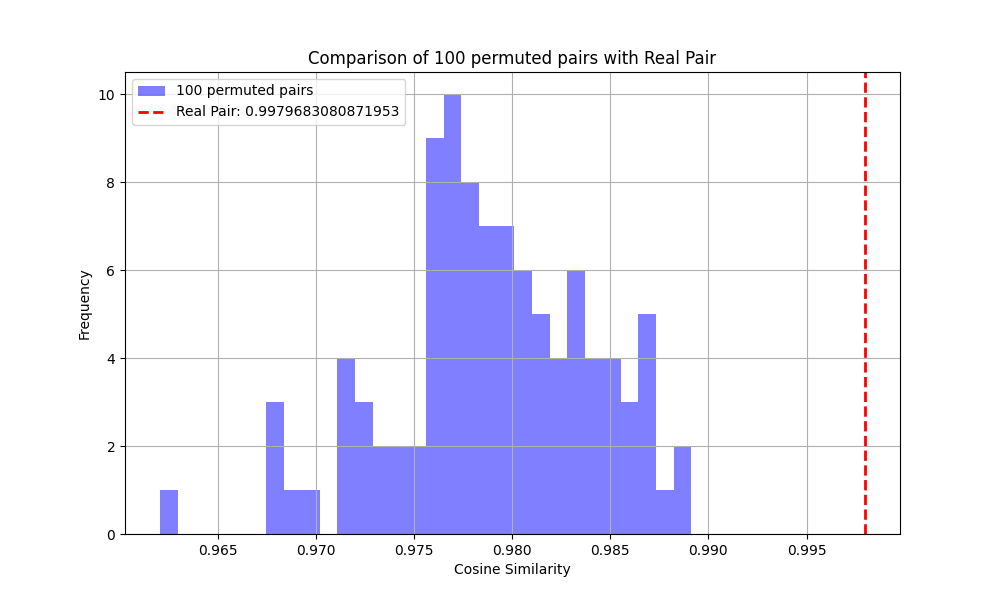}
        \caption{Cosine Similarity}
    \end{subfigure}
    \begin{subfigure}[H]{0.32\textwidth}
        \centering
        \includegraphics[width=\textwidth]{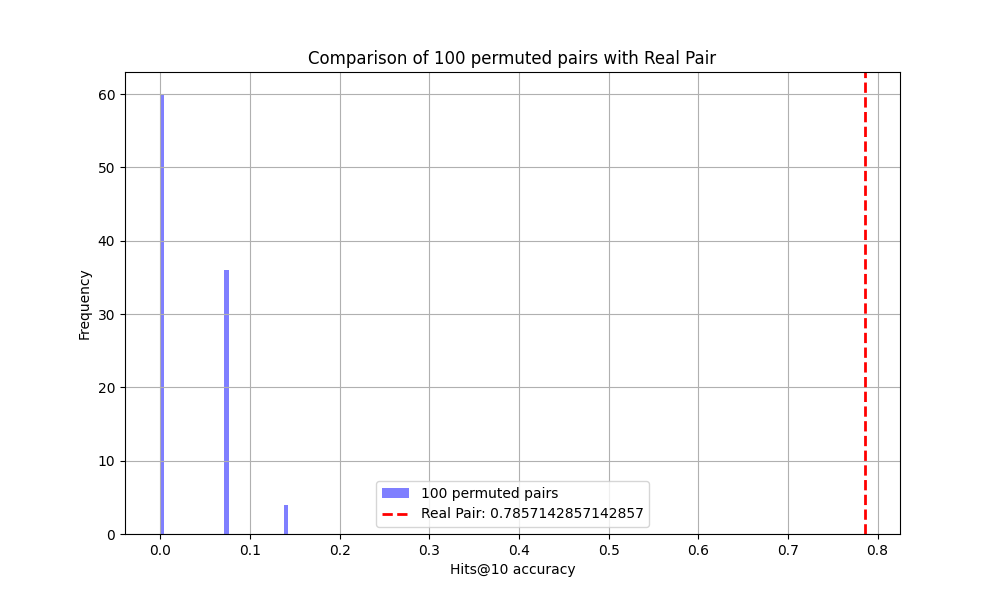}
        \caption{Retrieval Accuracy@1}
    \end{subfigure}
    \caption{The test statistics for user embedding decomposition. Dash line is the average performance of $ \hat{\textbf{B}} $ learned from the user embedding. The bars are the distribution of the results from random permutations that run for 100 times. }
    \label{fig:kg_results}
\end{figure*}

\begin{table*}
\centering
    \caption{p-value for hypothesis test. Note that * indicates better than random baseline to significance level $\alpha$ = 0.01. In our case, we are estimating directly the p-value, as the probability of an event, that we could have a high (low) value of the test-statistic by chance under the null-hypothesis}
    \begin{tabular}{@{}llllll@{}}
    \hline
                 & L2 Norm & Cosine Similarity & Retrieval Acc. & p-value \\ \hline
    Gender, Age Real Pair   & 0.47* & 99.8\%   & 0.79* & <0.01  \\ \hline
    Gender, Age Permuted    & 1.11-2.11*  & 96.5\%-99.0\%  & 0.00-0.14* &  <0.01  \\ \hline
    Gender, Age, Occ Real Pair& 17.87* &  97.1\%   &    0.23* &  <0.01 \\ \hline
    Gender, Age, Occ Permuted & 18.90-19.56* & 96.2\%-96.8\%  &    0.00-0.08* &  <0.01 \\ \hline
    \end{tabular}
    \label{tab: statistic test}
\end{table*}

\subsubsection{Additive Fusion Detection on Gender, Age and Occupation}
We afterwards group the user by gender, age and occupation and compute the mean embedding of 241 group of users.

\begin{figure*}[!h]
    \centering
    \begin{subfigure}[H]{0.32\textwidth}
        \centering
        \includegraphics[width=\textwidth]{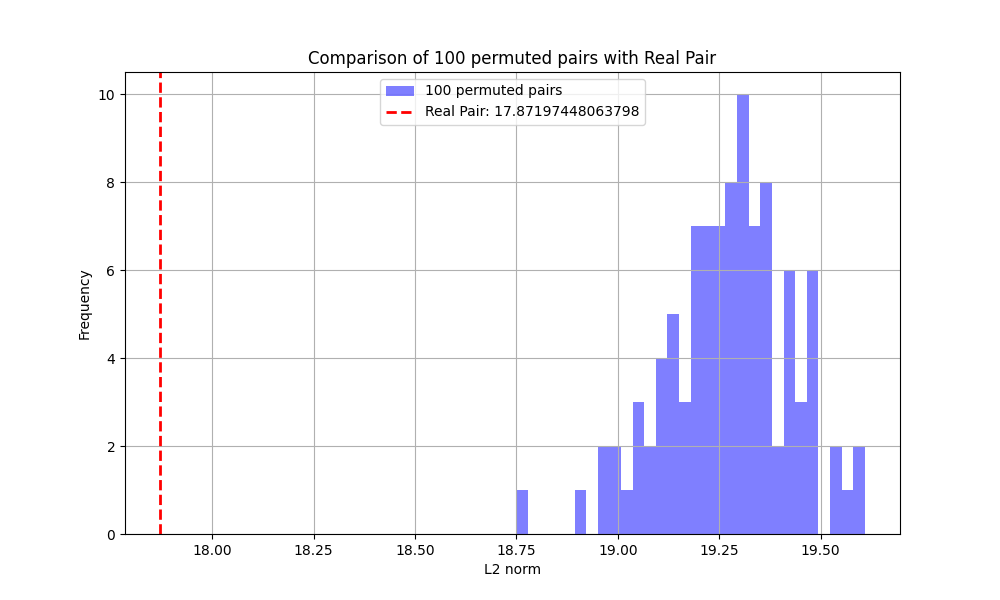}
        \centering
        \caption{Linear System Loss}
    \end{subfigure}
    \begin{subfigure}[H]{0.32\textwidth}
        \centering
        \includegraphics[width=\textwidth]{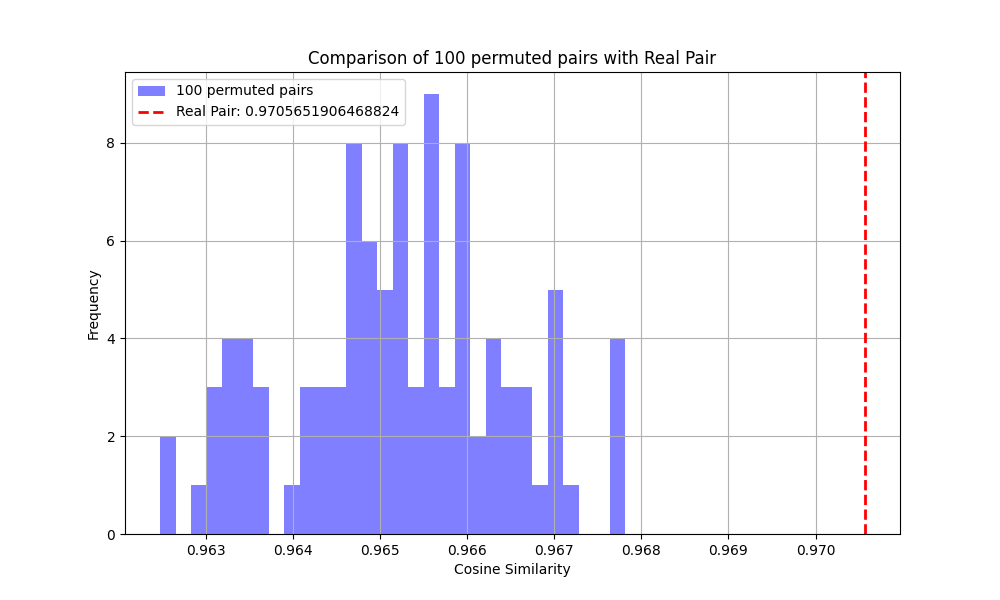}
        \caption{Cosine Similarity}
    \end{subfigure}
    \begin{subfigure}[H]{0.32\textwidth}
        \centering
        \includegraphics[width=\textwidth]{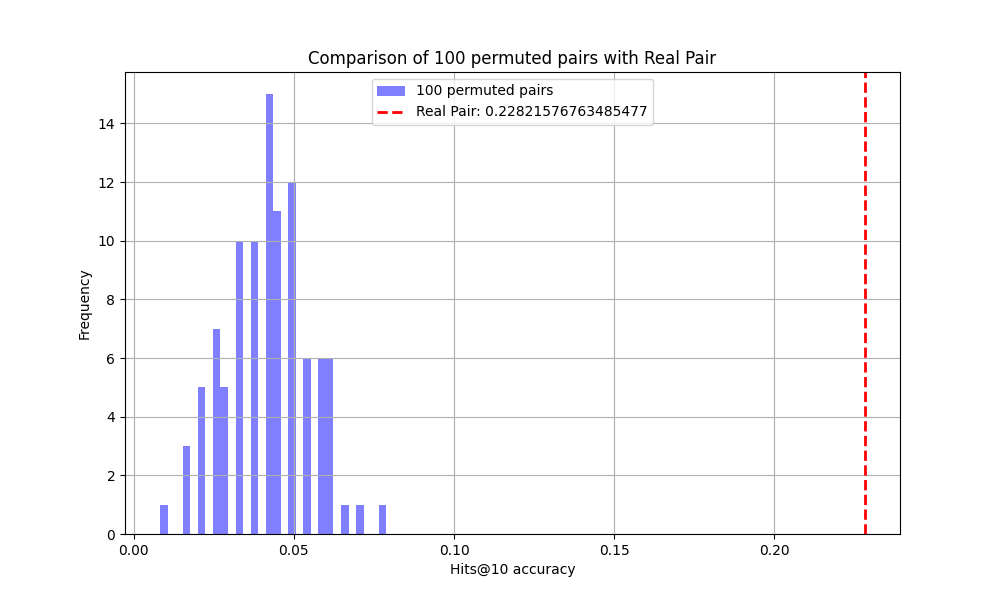}
        \caption{Retrieval Accuracy@10}
    \end{subfigure}
    \caption{The test statistics for user embedding decomposition. Dash line is the average performance of $ \hat{\textbf{B}} $ learned from the user embedding. The bars are the distribution of the results from random permutations that run for 100 times.}
    \label{fig:kg_occ_results}
\end{figure*}

When decomposing the embedding into gender, age and occupation, the L2 norm is 17.87 which is lower than every loss for a random permutation (18.90-19.56). As for identity retrieval accuracy, although the value is only 0.23 which is not a good result, it is still higher than any random permuted pairs (0.00-0.08). Therefore, the null hypothesis is rejected. Detailed information is shown in Figure \ref{fig:kg_occ_results}.
\section{Discussion}
\label{sec: discussion}
We have presented two methods for signals of compositionality detection in three different data types, word embedding, sentence embedding and graph embedding. 
\paragraph{Word Embedding}
Word2Vec's ability to capture deep semantic meanings becomes evident when compared with structured resources like WordNet. Even though Word2Vec operates in a continuous vector space, it surprisingly aligns well with these semantically organized databases. But its capabilities don't stop at semantics. When analyzed alongside tools like MorphoLex, it's clear that Word2Vec also grasps the subtle details of word formation, from roots to suffixes. These observations emphasize the depth of information embedded within word contexts — they don't just convey basic meaning, but also carry detailed linguistic information, including morphology. This richness within Word2Vec offers opportunities for in-depth analyses and insights into the multiple signals it derives from word context.

The diverse signals captured by Word2Vec lend it a structural richness that facilitates its decomposition. This has transformative implications. By segregating embeddings into distinct components, such as roots and suffixes, we can not only predict embeddings for novel words but also attain a granular understanding of the internal vector makeup. This dissection reaffirms that word contexts during training weave a multidimensional tapestry, intertwining semantics with morphology and more.
\paragraph{Sentence Embedding}
To examine the properties of sentence embedding, we have generated an SVO sentence corpus and embedded it with BERT. By applying a linear system, it has shown that the BERT sentence embedding can be decomposed into word representation with a linear system so that $\Phi_{BERT}(I_{i,j,k}) \approx \Phi_{LINEAR}(Sbj_{i}) + \Phi_{LINEAR}(Veb_{j}) + \Phi_{LINEAR}(Obj_{k})$. This allows for inference of a sentence embedding with simple linear algebra. The inference can have 77\% cosine similarity compared to the BERT sentence embedding. The learned word representation can also predict the embedding without seeing the sentence and achieve 64\% similarity. The results have shown that the BERT sentence embedding is compositional. However, it contains more properties than words and needs further analysis in future work.

\paragraph{Graph Embedding} we found that certain dimensions of user embeddings that relate to specific information should correlate with certain patterns of demographic information corresponding to the same meaning, across all users. Using the private attributes representation obtained in this way we first demonstrate that the correlations detected between the two versions of the user representation are significantly higher than random, and hence that a representation based on such features does capture statistical patterns that reflect private attribute information.

As for the linear system, we assume that user-behaviour-embedding is (approximated by) a sum of user-demographic vectors, showing that user embeddings can be decomposed into a weighted sum of attribute embeddings.
This refers to the compositionality of the user embedding, for example, the embedding of a ``50 year old female'' can be computed by the sum of the embedding of ``50'' and ``female''.
We can detect private attributes from both user embeddings in the movie system.

\section{Conclusions}
Three different types of data, word embedding, sentence embedding and knowledge graph embedding, present some compositionality, that is some of the information contained in them can be explained in terms of known attributes. This creates the possibility to manipulate those representations, for the purpose of removing bias, or to explain the decisions of the algorithm using them, or to answer analogical or counterfactual questions. 

In the case of word embedding, both the semantic and morphological information signals are detected from the context-based embedding. Sentence embedding, produced by BERT, presents some compositionality in terms of subject, verb, and object. In the case of movie recommender system, computed by the movie preference only, user embedding presents some compositionality of their private attributes such as age, gender and occupation. This creates the possibility to manipulate those representations, for the purpose of removing bias, or to explain the decisions of the algorithm using them, or to answer analogical or counterfactual questions.

\newpage
\bibliographystyle{cas-model2-names}

\bibliography{cas-refs}

\end{document}